\documentclass[10pt,twocolumn,letterpaper]{article}
\PassOptionsToPackage{table}{xcolor}

\usepackage[pagenumbers]{cvpr} 

\usepackage{booktabs}
\usepackage{makecell}
\usepackage{tabularx}
\usepackage{subcaption}
\usepackage{algorithm}
\usepackage{algpseudocode}
\usepackage{bibunits}
\definecolor{controlpink}{RGB}{250,224,232}
\definecolor{controlblue}{RGB}{220,234,248}
\newcommand{\pinkcell}[1]{\cellcolor{controlpink}#1}
\newcommand{\bluecell}[1]{\cellcolor{controlblue}#1}

\newcommand{\controlshade}[4]{%
  \pgfmathsetmacro{\controldelta}{#4*(#1-#2)}%
  \pgfmathparse{45*min(1,abs(\controldelta)/#3)}%
  \ifdim\controldelta pt>0pt
    \edef\controlcolor{%
      \noexpand\cellcolor{hardcolor!\pgfmathresult}}%
  \else
    \edef\controlcolor{%
      \noexpand\cellcolor{easycolor!\pgfmathresult}}%
  \fi
  \controlcolor #1%
}

\usepackage{graphicx}
\usepackage{booktabs}
\usepackage{multirow}
\usepackage[table]{xcolor}
\usepackage{pgfmath}

\definecolor{heatblue}{RGB}{65,125,180}

\newcommand{\heatcell}[3]{%
  \pgfmathparse{8 + 42 * max(0,min(1,(#1-#2)/(#3-#2)))}%
  \edef\heatcommand{%
    \noexpand\cellcolor{heatblue!\pgfmathresult}%
  }%
  \heatcommand #1%
}

\usepackage{graphicx}
\usepackage[table]{xcolor}

\definecolor{attractcolor}{RGB}{230,145,56}
\definecolor{avoidcolor}{RGB}{70,130,180}

\definecolor{cvprblue}{rgb}{0.21,0.49,0.74}
\usepackage[pagebackref,breaklinks,colorlinks,allcolors=cvprblue]{hyperref}
\definecolor{oursgray}{gray}{0.92}
\def\confName{CVPR}
\def\confYear{2027}
\usepackage{microtype}

\title{Controllable Crowd Generation through World-Model Planning\vspace{-4mm}}

\author{JunGyu Lee$^{1}$ \quad Jisu Shin$^{2}$ \quad Seunghyun Shin$^{2}$ \quad Hae-Gon Jeon$^{1,*}$\\
$^{1}$Yonsei University \quad $^{2}$GIST\\[-0.2em]
{\small\href{https://jungyu0413.github.io/Ctrl-CWM}{\texttt{https://jungyu0413.github.io/Ctrl-CWM}}}}

\begin{document}

\twocolumn[{%
\renewcommand\twocolumn[1][]{#1}%
\maketitle
\begin{center}
\vspace{-9mm}
\includegraphics[width=\textwidth]{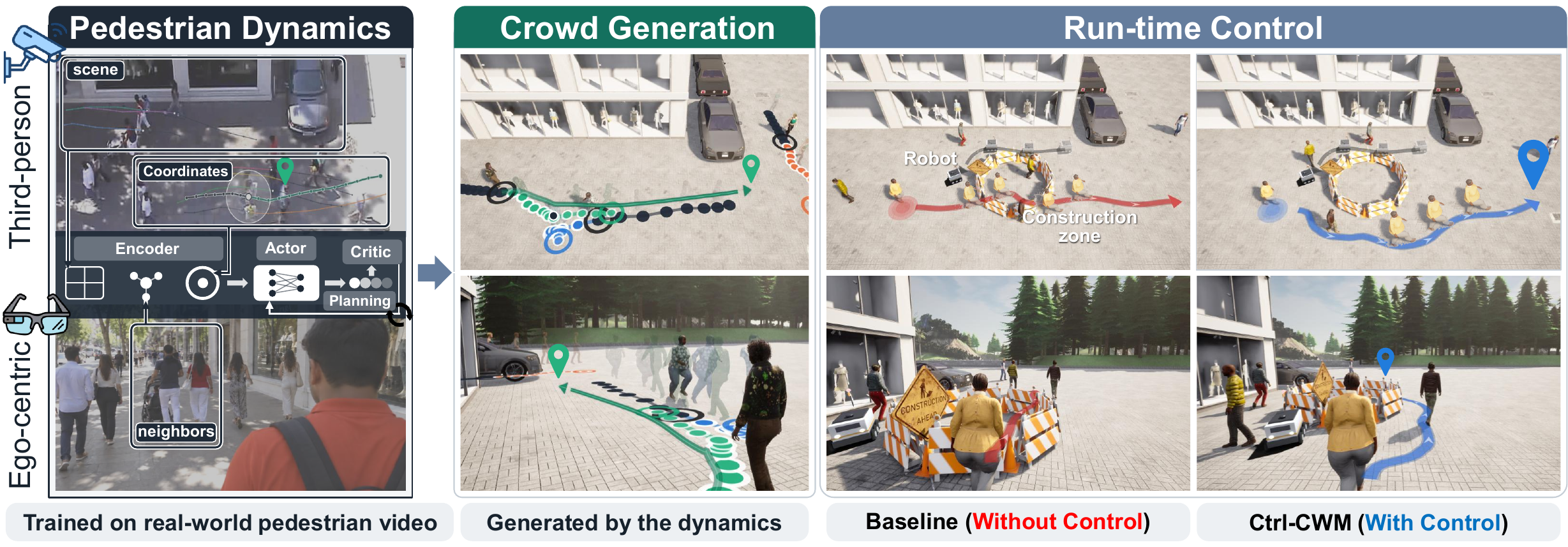}

\vspace{-3mm}

\captionsetup{hypcap=false}
\captionof{figure}{\textbf{Ctrl-CWM.}
Human dynamics learned from real-world pedestrian video
\textbf{(left)} are used for continuous crowd generation
\textbf{(center)} and runtime crowd control \textbf{(right)}.
Characters are rendered in CARLA on the generated trajectories.}
\label{fig:teaser}
\end{center}

}]
{\renewcommand{\thefootnote}{}\footnotetext{$^{*}$Corresponding author.}}

\begin{abstract}
When a robot approaches or an obstacle blocks the way, pedestrians may change their paths in real time. Therefore, crowd simulators should reproduce realistic human motion and allow pedestrian paths to be changed during simulation. However, existing methods offer limited flexibility for run-time crowd control. In this paper, we propose \textbf{Ctrl-CWM}, a multi-agent \textbf{C}on\textbf{t}\textbf{r}o\textbf{l}lable \textbf{C}rowd \textbf{W}orld \textbf{M}odel that integrates continuous crowd trajectory generation with run-time control. Our key idea is to combine world-model planning with a motion representation learned by predicting pedestrian trajectories annotated in real-world videos. We simulate crowd movements under candidate actions, evaluate each pedestrian's motion with a learned critic, and select actions to guide pedestrians away from obstacles or toward target locations. Users can update these control objectives during simulation without retraining. We assess the crowd generation in various pedestrian entry patterns and the run-time control through avoidance and attraction tasks. Ctrl-CWM outperforms baseline simulators on most generation metrics and achieves higher control compliance. Furthermore, in robot navigation simulations with three fixed navigation methods, Ctrl-CWM consistently produces the socially acceptable pedestrian responses according to the robot's motions.

\vspace{-3mm}
\end{abstract}

\section{Introduction}
\label{sec:intro}

Crowd simulation supports robot navigation, autonomous driving, and urban planning by reproducing how pedestrians move and respond according to environmental changes. For example, pedestrians may avoid an approaching robot or vehicle, or take a different route when a walkway becomes blocked. To capture these responses, simulators should generate realistic pedestrian motion and allow users to change pedestrians' paths during simulation.

Traditional crowd simulators~\cite{reynolds1987flocks,helbing1995social,van2010optimal} model pedestrian navigation and interactions using hand-designed rules. Learning-based approaches capture human behaviors from pedestrian data or simulated experiences to generate realistic crowds~\cite{lee2018crowd,panayiotou2022ccp,charalambous2023greil}. CrowdES~\cite{bae2025crowdes} generates continuous crowds from single scene images. 

However, these methods rely on predefined behaviors and parameters for control, limiting changes in walking patterns. Diffusion-based approaches~\cite{rempe2023trace} support flexible controls at high computational costs that can limit real-time use. This motivates us to develop efficient planning for run-time control with socially acceptable motion patterns of agents.

Recent advances in world models~\cite{hafner2025mastering,hansen2022temporal,hansen2024td} have shown how simulating futures can support behavior learning and planning. However, adapting world models to crowd simulation requires anticipating how pedestrians respond to others while maintaining realistic motion as control objectives change. Generating realistic crowd motion with world models therefore remains challenging.

We propose Ctrl-CWM, a multi-agent Controllable Crowd World Model that simultaneously supports  continuous crowd generation and run-time control (Fig.~\ref{fig:teaser}), consisting of a crowd representation module, an actor, a critic and a sampling-based planner. The actor, critic and planner follow the world-model planning principle. For crowd generation, the actor simulates how surrounding pedestrians respond to candidate actions, and the critic evaluates each pedestrian's motion quality within these simulations. The planner utilizes these evaluations to select movements that guide pedestrians toward or away from locations specified by users.

We first pre-train a trajectory predictor on trajectories from real-world videos to learn a crowd representation of scene context and pedestrian interactions. We freeze its feature extractor and train the actor to infer each pedestrian's next movement from updated crowd states. The critic learns to assign higher scores to realistic candidate trajectories.

At run time, users specify areas for agents to avoid or locations to go near. The planner selects agents' movements based on the critic's motion scores and the user's objective. We plan their next movements from the pedestrians' updated positions. Here, we make use of the user objectives only during planning, allowing them to change without retraining.

We evaluate the crowd generation on public benchmark datasets such as ETH--UCY, SDD, and GCS using scene-level and agent-level metrics. The control experiments introduce avoidance and attraction objectives during simulation. Ctrl-CWM achieves the best average scores on most generation metrics and high compliance with control objectives. To assess crowd control in downstream robot navigation, we exploit Habitat 3.0~\cite{puig2024habitat} with three fixed navigation methods. 
In this downstream task, we validate that the crowd generated by Ctrl-CWM responds to moving robots and can improve the navigation success and reduce the collision rates compared with the no-objective conditions in unseen indoor environments without re-training.
\section{Related Work}
We review three areas relevant to Ctrl-CWM: trajectory prediction for learning pedestrian motion representations, crowd simulation for generation and control, and world-model planning for evaluating actions through imagined futures.
\subsection{Human Trajectory Prediction} 
Human trajectory prediction models future pedestrian motion from observed trajectories and scene context~\cite{alahi2016social,gupta2018social}. Interactions are represented through graphs~\cite{mohamed2020social,shi2021sgcn}, groups~\cite{zhou2022grouptron}, and agent-aware attention~\cite{yuan2021agentformer}. Scene features and destination estimates further condition predictions~\cite{sadeghian2019sophie,mangalam2021goals}. Multimodal forecasting uses conditional VAEs~\cite{salzmann2020trajectron++}, diffusion models~\cite{gu2022stochastic}, flow-matching models~\cite{fu2025moflow}, and language models~\cite{bae2024can,kim2025guide}. We use trajectory prediction to pretrain motion and interaction representations for closed-loop crowd generation. The pretrained feature extractor remains fixed during simulator training.

\begin{figure*}[t]
  \centering
  \includegraphics[width=\textwidth]{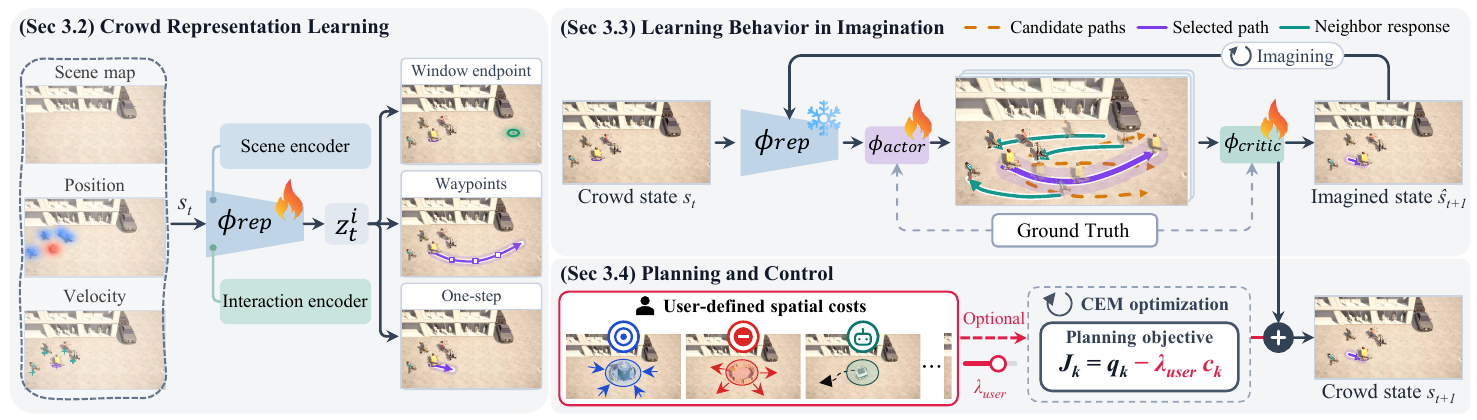}
  \vspace{-6mm}
  \caption{\textbf{Ctrl-CWM.}
  Trajectory prediction pretrains the crowd representation
  module $\phi_{\mathrm{rep}}$, which is then frozen.
  The actor $\phi_{\mathrm{actor}}$ generates imagined crowd
  trajectories through repeated explicit state updates.
  The critic $\phi_{\mathrm{critic}}$ learns to rank candidate
  rollouts using trajectory-derived supervision.
  The planner combines critic scores with user-defined spatial costs
for run-time control.}
  \label{fig:overview}
  \vspace{-3mm}
\end{figure*}

\subsection{Crowd Simulation and Control}
Rule-based simulators use flocking~\cite{reynolds1987flocks}, social forces~\cite{helbing1995social}, and reciprocal collision avoidance~\cite{van2010optimal}, without learning motion patterns from data. Continuum approaches combine navigation with collective flow~\cite{treuille2006continuum}, while example-based systems select or blend recorded motions~\cite{lee2007group,lerner2007crowds,ju2010morphable}. User control has been explored through navigation fields derived from sketches or video~\cite{patil2010directing}, proxy agents that influence surrounding pedestrians~\cite{yeh2008composite}, and sketched velocity and orientation fields~\cite{colas2022interaction}. Interactive editing allows users to modify crowd animations~\cite{kim2014interactive}.

Reinforcement-learning approaches model pedestrian navigation~\cite{lee2018crowd}. CCP~\cite{panayiotou2022ccp} conditions policies on behavior weights adjustable at run time. GREIL-Crowds~\cite{charalambous2023greil} learns behaviors using rewards derived from observed crowds. CEDRL~\cite{panayiotou2025cedrl} learns from multiple datasets and conditions its policy on behavioral complexity. These methods focus on pedestrian navigation within simulation environments rather than generating crowds directly from an input scene image. TRACE~\cite{rempe2023trace} generates trajectories conditioned on scene images and supports run-time control through diffusion guidance and replanning, but iterative denoising can make frequent replanning computationally expensive. Diffusion models support social-physics-informed crowd simulation~\cite{chen2024social} and text-guided generation of crowd distribution and velocity fields~\cite{ji2024text}. CrowdES~\cite{bae2025crowdes} generates continuous crowds from scene images, but its parameter-level controls offer limited flexibility for adapting ongoing trajectories to new avoidance or attraction objectives during simulation.

Following CrowdES, we study continuous crowd generation conditioned on real-world scene images. We adopt its scene-level and agent-level evaluation protocol and focus on run-time avoidance and attraction objectives. Ctrl-CWM integrates a learned motion critic with sampling-based planning to generate trajectories according to user objectives, which can change during simulation without retraining.

\subsection{World Models and Planning}
World models support behavior learning and planning by predicting the consequences of actions~\cite{ha2018world,hafner2025mastering}. Dreamer learns policies from trajectories imagined in a learned latent dynamics model~\cite{hafner2019dream}. TD-MPC methods combine learned latent dynamics with value estimates for online planning~\cite{hansen2022temporal,hansen2024td}, while PETS uses probabilistic dynamics ensembles to plan in the observed state space~\cite{chua2018deep}. Ctrl-CWM adapts world-model planning to continuous multi-agent crowd generation by simulating how surrounding pedestrians respond to candidate actions. A learned critic scores each pedestrian's motion within these simulations for action selection.
\section{Method}

\subsection{Problem Formulation}
\label{sec:formulation}
As shown in \cref{fig:overview}, Ctrl-CWM consists of a crowd representation module, an actor, a critic, and a sampling-based planner. The representation module encodes scene context and pedestrian interactions (\cref{sec:worldmodel}). The actor proposes next-step displacements from the representation, while the critic scores the resulting imagined trajectories (\cref{sec:stage2}). The planner then selects each controlled pedestrian's next-step displacement based on these scores and user-defined spatial costs, such as avoiding hazardous regions or approaching designated targets (\cref{sec:planning}).

At time $t$, the crowd state $s_t$ contains the static scene map and, for each active pedestrian $i$, its position $p_t^i$, velocity history, and local goal. Let $v_t^i$ denote its current velocity and $u_t^i$ the unit direction toward its local goal. We denote an imagined crowd sequence over a planning horizon of $H$ steps by $\hat s_{t+1:t+H}$. The three modules operate as follows:
\begin{equation}
\begin{aligned}
\text{Crowd representation}\qquad z_t^i
    &= \phi_{\mathrm{rep}}(s_t, i), \\
\text{Actor}\qquad \bar a_t^i
    &= \phi_{\mathrm{actor}}(z_t^i, u_t^i, v_t^i), \\
\text{Critic}\qquad q_t^i
    &= \phi_{\mathrm{critic}}(\hat s_{t+1:t+H}, i).
\end{aligned}
\label{eq:components}
\end{equation}
Here, $z_t^i$ represents scene context, recent motion, and interactions with neighbors from pedestrian $i$'s perspective, and $\bar a_t^i$ is its proposed next-step displacement. The critic's motion score $q_t^i$ evaluates pedestrian $i$'s motion within $\hat s_{t+1:t+H}$. All three modules are shared across pedestrians.

For a candidate rollout, we replace pedestrian $i$'s first displacement with a candidate displacement while retaining the actor proposals for the others. At subsequent steps, the actor recomputes displacement proposals for all pedestrians from the updated crowd state. Positions and velocity histories are updated each step while the scene map remains fixed.

For pedestrian $i$, $a_t^i$ denotes the displacement selected by the planner. All active pedestrian positions are updated simultaneously:
\begin{equation}
p_{t+1}^i =
\begin{cases}
p_t^i + a_t^i,      & \text{if selected for planning}, \\
p_t^i + \bar a_t^i, & \text{otherwise}.
\end{cases}
\label{eq:transition}
\end{equation}
We update velocity histories from the resulting motion and replan at the next simulation step.

\subsection{Crowd Representation Learning}
\label{sec:worldmodel}

We pre-train a trajectory prediction network on annotated pedestrian trajectories from real-world videos to capture scene context and motion patterns. Its feature extractor becomes the crowd representation module $\phi_{\mathrm{rep}}$.

\paragraph{Scene and interaction encoding.}
The scene encoder uses a U-Net~\cite{ronneberger2015u} to capture spatial relationships between pedestrian locations and the surrounding environment. Its inputs consist of the semantic scene map and separate heatmaps for pedestrian $i$ and the crowd, aligned with the map. The interaction encoder complements these features with recent motion and relations to nearby pedestrians. It combines a GRU encoding of pedestrian $i$'s velocity history~\cite{cho2014learning} with attention over neighbors based on relative positions. We integrate the scene and interaction features to obtain $z_t^i$.

\noindent\textbf{Trajectory supervision.}
We supervise the representation over an $L$-step window with three prediction heads for a pedestrian's future endpoint, path, and next-step motion. For a given pedestrian at time $t$, let $p_j^*$ denote its ground-truth position at $j$ steps after time  $t$, for $j=1,\ldots,L$. The goal head infers a heatmap for the window endpoint $p_L^*$, while the waypoint head predicts heatmaps for the sequence $p_1^*,\ldots,p_L^*$. An auxiliary head provides a heatmap for $p_1^*$ to emphasize the immediate motion needed for its future displacement prediction. During pre-training, the waypoint and auxiliary heads are conditioned on the ground-truth endpoint. We jointly train the feature extractor and prediction heads using a linear combination of the three prediction losses as follows:
\begin{equation}
\mathcal L_{\mathrm{crowd}}
= \mathcal L_{\mathrm{goal}}
+ \lambda_{\mathrm{way}}\mathcal L_{\mathrm{way}}
+ \lambda_{\mathrm{aux}}\mathcal L_{\mathrm{aux}} .
\label{eq:method_pred_summary}
\end{equation}
All three heads use weighted binary cross-entropy against Gaussian heatmaps centered at the ground-truth positions. The losses $\mathcal{L}_{\mathrm{goal}}$ and $\mathcal{L}_{\mathrm{way}}$ additionally penalize Euclidean distances between soft-argmax coordinates and ground-truth positions. $\mathcal L_{\mathrm{way}}$ is averaged over the $L$ steps. Details of the loss functions and weights are provided in Appendix A.2. After pretraining, $\phi_{\mathrm{rep}}$ remains frozen during training.

\subsection{Learning Behavior in Imagination}
\label{sec:stage2}
We train the actor and critic on imagined crowd rollouts. The actor learns to generate motion consistent with ground-truth trajectories, while the critic learns to rank candidate first-step displacements by evaluating the crowd motion that follows.

\noindent\textbf{Actor.}
We train the actor $\phi_{\mathrm{actor}}$ to generate $L$-step crowd trajectories that reproduce the motion patterns of real pedestrians. Starting from an observed crowd state $s_t$, we assign each pedestrian a nearby future position sampled from its ground-truth trajectory as its training goal. At rollout step $h=0,\ldots,L-1$, $\phi_{\mathrm{actor}}$ predicts the displacement $\bar a_{t+h}^i$ for each pedestrian. We apply these displacements simultaneously to update positions and velocity histories, then recompute $z_{t+h+1}^i$ from the updated crowd state for the next prediction step. We optimize the actor as follows:
\begin{equation}
\mathcal L_{\mathrm{act}}
= \mathcal L_{\mathrm{imit}}
+ \lambda_{\mathrm{kin}}^{\mathrm a}\mathcal L_{\mathrm{kin}}
+ \lambda_{\mathrm{adv}}\mathcal L_{\mathrm{adv}}.
\label{eq:method_actor_summary}
\end{equation}
The \emph{trajectory imitation loss} $\mathcal L_{\mathrm{imit}}$ combines a stepwise position error with soft-DTW~\cite{cuturi2017soft}, which compares generated and ground-truth paths while allowing temporal misalignment. The \emph{kinematic loss} $\mathcal L_{\mathrm{kin}}$ matches generated speeds and their step-to-step changes to those in the ground-truth trajectories. The least-squares \emph{adversarial loss} $\mathcal L_{\mathrm{adv}}$~\cite{mao2017least} trains the actor to generate displacement sequences that a discriminator classifies as real motion. We calibrate $\lambda_{\mathrm{kin}}^{\mathrm a}$ using the relative gradient norms of the actor losses, while $\lambda_{\mathrm{adv}}$ controls the strength of adversarial supervision.

\noindent\textbf{Critic.}
We train the critic $\phi_{\mathrm{critic}}$ to score candidate displacements using supervision from real pedestrian trajectories. For an observed crowd state $s_t$ and pedestrian $i$, we sample $K$ alternative displacements around the actor proposal $\bar a_t^i$. We simulate each alternative for $L$ steps as described in~\cref{sec:formulation}. At the first step, pedestrian $i$ uses the sampled displacement, while the others use the actor's predictions. At subsequent steps, the actor infers all pedestrians' displacements from the updated crowd state. For candidate $k$, we compute the following trajectory cost for pedestrian $i$:
\begin{equation}
y_k = \mathcal L_{\mathrm{imit}}
+ \lambda_{\mathrm{kin}}^{\mathrm c}\mathcal L_{\mathrm{kin}}
+ \lambda_{\mathrm{col}}\mathcal L_{\mathrm{col}}.
\label{eq:long_cost}
\end{equation}
We use the imitation and kinematic terms from actor training (Eq.~\ref{eq:method_actor_summary}), comparing each candidate's simulated trajectory for pedestrian $i$ with the same ground-truth trajectory. The \emph{collision term} $\mathcal L_{\mathrm{col}}$ counts neighboring pedestrians within a fixed radius of pedestrian $i$ and averages the count over the $L$ predicted steps. We calibrate $\lambda_{\mathrm{kin}}^{\mathrm c}$ and $\lambda_{\mathrm{col}}$ using the relative standard deviations of the loss terms across candidates.

For candidate $k$, the critic provides a motion score $q_k=\phi_{\mathrm{critic}}(\hat s_{t+1:t+H}^{(k)},i)$ from only the first $H<L$ steps of its simulated crowd trajectory. Its inputs encode pedestrian $i$'s motion and geometric relationships with neighboring pedestrians and the local goal. We supervise these scores with the $L$-step costs using a listwise cross-entropy loss~\cite{cao2007learning}:
\begin{equation}
\begin{gathered}
\mathcal L_{\mathrm{critic}}
    = -\sum_{k=1}^{K} P_k \log Q_k \quad \text{s.t.}\\ 
P_k = \operatorname{softmax}(-\mathbf y/\tau_c)_k,
\quad
Q_k = \operatorname{softmax}(\mathbf q/\tau_c)_k ,
\end{gathered}
\label{eq:listnet}
\end{equation}
where $\mathbf y=(y_1,\ldots,y_K)$ contains the training costs, $\mathbf q=(q_1,\ldots,q_K)$ denotes the predicted motion scores for the same state and pedestrian, and $\tau_c>0$ is the softmax temperature. Matching $\mathbf Q$ to $\mathbf P$ encourages the critic to assign higher scores to lower-cost candidates. Full actor and critic training details are provided in Appendix A.2.

\begin{table*}[t]
\vspace{-1mm}
\caption{\textbf{Crowd behavior generation under shared arrivals.}
Results use identical replayed random-surface (left) and diffusion (right) arrivals across methods.
AVG averages five ETH--UCY folds, SDD, and GCS. Col.\ is in \%.
\textbf{Bold}/\underline{underline}: best/second-best.}
  \vspace{-2mm}
  \label{tab:gen_shared}
  \centering
  \setlength{\tabcolsep}{2pt}
  \resizebox{\textwidth}{!}{%
    \begin{tabular}{llcccccccc}
      \toprule
      \multicolumn{10}{c}{\textit{Arrivals from the RS emitter}} \\
      \midrule
      \multirow{2}{*}{Dataset/Fold} & \multirow{2}{*}{Model}
        & \multicolumn{4}{c}{Scene-Level Realism}
        & \multicolumn{4}{c}{Agent-Level Accuracy} \\
      \cmidrule(lr){3-6} \cmidrule(lr){7-10}
        & & Dens.$\downarrow$ & Freq.$\downarrow$ & Cov.$\downarrow$ & Pop.$\downarrow$ & Kinem.$\downarrow$ & DTW$\downarrow$ & Div.$\uparrow$ & Col.(\%)$\downarrow$ \\
      \midrule
      \multirow{4}{*}{ETH} & ORCA & \textbf{0.019} & \underline{0.016} & \underline{0.016} & \textbf{0.205} & 0.839 & 2.848 & 0.092 & \textbf{0.004} \\
 & TRACE & \underline{0.028} & \textbf{0.010} & \textbf{0.010} & \underline{0.329} & \underline{0.381} & \underline{1.666} & \underline{0.179} & 0.335 \\
 & CrowdES & 0.044 & 0.025 & 0.025 & 0.461 & 1.138 & 2.608 & \textbf{0.240} & 2.660 \\
\rowcolor{gray!20}
\cellcolor{white} & \textbf{Ctrl-CWM} & 0.034 & \underline{0.016} & \underline{0.016} & 0.398 & \textbf{0.358} & \textbf{1.511} & 0.171 & \underline{0.265} \\
      \midrule
      \multirow{4}{*}{HOTEL} & ORCA & 0.044 & 0.017 & 0.017 & 0.396 & 0.623 & 1.094 & 0.179 & \textbf{0.015} \\
 & TRACE & 0.032 & 0.029 & 0.029 & 0.266 & 0.566 & \textbf{0.722} & \underline{0.219} & 1.118 \\
 & CrowdES & \underline{0.015} & \underline{0.016} & \underline{0.016} & \underline{0.130} & \underline{0.524} & 0.907 & 0.184 & 2.471 \\
\rowcolor{gray!20}
\cellcolor{white} & \textbf{Ctrl-CWM} & \textbf{0.012} & \textbf{0.012} & \textbf{0.012} & \textbf{0.098} & \textbf{0.419} & \underline{0.746} & \textbf{0.224} & \underline{0.828} \\
      \midrule
      \multirow{4}{*}{UNIV} & ORCA & \textbf{0.311} & \textbf{0.210} & \textbf{0.210} & \textbf{0.658} & \textbf{0.357} & 1.878 & 0.221 & \textbf{0.007} \\
 & TRACE & 0.363 & \underline{0.216} & \underline{0.216} & 0.769 & 0.627 & \underline{1.235} & \textbf{0.338} & 0.919 \\
 & CrowdES & 0.365 & 0.230 & 0.230 & 0.771 & \underline{0.457} & 1.394 & 0.249 & 2.758 \\
\rowcolor{gray!20}
\cellcolor{white} & \textbf{Ctrl-CWM} & \underline{0.362} & \underline{0.216} & \underline{0.216} & \underline{0.765} & 0.543 & \textbf{1.184} & \underline{0.327} & \underline{0.733} \\
      \midrule
      \multirow{4}{*}{ZARA1} & ORCA & 0.063 & 0.027 & 0.027 & 0.953 & 1.435 & 2.530 & 0.181 & \textbf{0.002} \\
 & TRACE & 0.016 & 0.010 & 0.010 & 0.209 & 0.773 & 1.145 & 0.216 & \underline{0.585} \\
 & CrowdES & \underline{0.011} & \underline{0.009} & \underline{0.009} & \underline{0.167} & \textbf{0.358} & \underline{0.706} & \textbf{0.283} & 2.010 \\
\rowcolor{gray!20}
\cellcolor{white} & \textbf{Ctrl-CWM} & \textbf{0.010} & \textbf{0.008} & \textbf{0.008} & \textbf{0.137} & \underline{0.691} & \textbf{0.703} & \underline{0.258} & 1.015 \\
      \midrule
      \multirow{4}{*}{ZARA2} & ORCA & \underline{0.023} & \underline{0.010} & \underline{0.010} & \underline{0.247} & 0.742 & 1.358 & 0.110 & \textbf{0.015} \\
 & TRACE & 0.036 & 0.015 & 0.015 & 0.368 & 0.714 & 1.117 & 0.194 & \underline{0.669} \\
 & CrowdES & \textbf{0.020} & \textbf{0.008} & \textbf{0.008} & \textbf{0.191} & \textbf{0.340} & \underline{0.580} & \textbf{0.333} & 2.983 \\
\rowcolor{gray!20}
\cellcolor{white} & \textbf{Ctrl-CWM} & 0.026 & 0.012 & 0.012 & 0.258 & \underline{0.581} & \textbf{0.569} & \underline{0.306} & 1.457 \\
      \midrule
      \multirow{4}{*}{SDD} & ORCA & 0.061 & 0.045 & 0.043 & \underline{0.472} & \underline{0.684} & \textbf{5.168} & 0.360 & \textbf{0.011} \\
 & TRACE & 0.051 & 0.036 & 0.035 & 0.473 & 0.709 & 6.642 & \textbf{0.442} & \underline{0.031} \\
 & CrowdES & \textbf{0.043} & \underline{0.034} & \underline{0.033} & \textbf{0.458} & \textbf{0.662} & 6.303 & 0.375 & 1.643 \\
\rowcolor{gray!20}
\cellcolor{white} & \textbf{Ctrl-CWM} & \underline{0.046} & \textbf{0.033} & \textbf{0.031} & 0.476 & 0.718 & \underline{5.511} & \underline{0.383} & 0.227 \\
      \midrule
      \multirow{4}{*}{GCS} & ORCA & 6.787 & \textbf{0.369} & \textbf{0.369} & 15.683 & 2.423 & 5.120 & 0.232 & \textbf{0.011} \\
 & TRACE & \textbf{2.436} & 0.497 & 0.497 & \textbf{5.628} & \textbf{0.325} & \textbf{1.757} & \underline{0.338} & \underline{5.475} \\
 & CrowdES & 4.726 & 0.573 & 0.573 & 10.921 & 0.755 & 2.928 & \textbf{0.397} & 9.613 \\
\rowcolor{gray!20}
\cellcolor{white} & \textbf{Ctrl-CWM} & \underline{2.847} & \underline{0.371} & \underline{0.371} & \underline{6.578} & \underline{0.503} & \underline{1.766} & 0.327 & 7.645 \\
      \midrule
      \multirow{4}{*}{\textbf{AVG}} & ORCA & 1.044 & \underline{0.099} & \underline{0.099} & 2.659 & 1.015 & 2.857 & 0.196 & \textbf{0.009} \\
 & TRACE & \textbf{0.423} & 0.116 & 0.116 & \textbf{1.149} & \underline{0.585} & \underline{2.041} & 0.275 & \underline{1.305} \\
 & CrowdES & 0.746 & 0.128 & 0.128 & 1.871 & 0.605 & 2.204 & \textbf{0.294} & 3.448 \\
\rowcolor{gray!20}
\cellcolor{white} & \textbf{Ctrl-CWM} & \underline{0.477} & \textbf{0.095} & \textbf{0.095} & \underline{1.244} & \textbf{0.545} & \textbf{1.713} & \underline{0.285} & 1.739 \\
      \bottomrule
    \end{tabular}%
    \hspace{8pt}%
    \begin{tabular}{llcccccccc}
      \toprule
      \multicolumn{10}{c}{\textit{Arrivals from the diffusion emitter}} \\
      \midrule
      \multirow{2}{*}{Dataset/Fold} & \multirow{2}{*}{Model}
        & \multicolumn{4}{c}{Scene-Level Realism}
        & \multicolumn{4}{c}{Agent-Level Accuracy} \\
      \cmidrule(lr){3-6} \cmidrule(lr){7-10}
        & & Dens.$\downarrow$ & Freq.$\downarrow$ & Cov.$\downarrow$ & Pop.$\downarrow$ & Kinem.$\downarrow$ & DTW$\downarrow$ & Div.$\uparrow$ & Col.(\%)$\downarrow$ \\
      \midrule
      \multirow{4}{*}{ETH} & ORCA & 0.026 & 0.013 & 0.013 & 0.265 & 0.867 & 3.794 & 0.096 & \textbf{0.006} \\
 & TRACE & \textbf{0.013} & \underline{0.010} & \underline{0.010} & \textbf{0.142} & \underline{0.229} & 1.811 & 0.175 & \underline{0.469} \\
 & CrowdES & 0.020 & 0.020 & 0.020 & 0.208 & 0.377 & \underline{1.649} & \textbf{0.203} & 0.697 \\
\rowcolor{gray!20}
\cellcolor{white} & \textbf{Ctrl-CWM} & \underline{0.017} & \textbf{0.006} & \textbf{0.006} & \underline{0.193} & \textbf{0.214} & \textbf{1.458} & \underline{0.181} & 0.620 \\
      \midrule
      \multirow{4}{*}{HOTEL} & ORCA & 0.022 & \underline{0.010} & \underline{0.010} & 0.199 & 1.052 & 1.588 & 0.112 & \textbf{0.013} \\
 & TRACE & \underline{0.012} & \underline{0.010} & \underline{0.010} & \underline{0.092} & 0.341 & \underline{0.598} & \underline{0.256} & 1.517 \\
 & CrowdES & 0.013 & \textbf{0.009} & \textbf{0.009} & 0.117 & \underline{0.336} & 0.643 & 0.242 & 1.197 \\
\rowcolor{gray!20}
\cellcolor{white} & \textbf{Ctrl-CWM} & \textbf{0.011} & \textbf{0.009} & \textbf{0.009} & \textbf{0.087} & \textbf{0.305} & \textbf{0.548} & \textbf{0.259} & \underline{1.046} \\
      \midrule
      \multirow{4}{*}{UNIV} & ORCA & 0.348 & 0.214 & 0.214 & 0.738 & 0.732 & 1.884 & 0.221 & \textbf{0.014} \\
 & TRACE & 0.350 & 0.207 & 0.207 & 0.742 & 0.474 & \underline{1.136} & \textbf{0.354} & 1.317 \\
 & CrowdES & \underline{0.347} & \underline{0.204} & \underline{0.204} & \underline{0.734} & \textbf{0.420} & \textbf{1.121} & 0.340 & 0.645 \\
\rowcolor{gray!20}
\cellcolor{white} & \textbf{Ctrl-CWM} & \textbf{0.339} & \textbf{0.196} & \textbf{0.196} & \textbf{0.717} & \underline{0.436} & \textbf{1.121} & \underline{0.349} & \underline{0.609} \\
      \midrule
      \multirow{4}{*}{ZARA1} & ORCA & 0.064 & 0.024 & 0.024 & 0.974 & 1.950 & 2.594 & 0.181 & \textbf{0.002} \\
 & TRACE & \underline{0.012} & \underline{0.013} & \underline{0.013} & \underline{0.180} & \textbf{0.293} & 0.875 & \underline{0.285} & 1.242 \\
 & CrowdES & 0.018 & 0.017 & 0.017 & 0.254 & \underline{0.327} & \textbf{0.675} & \textbf{0.304} & 1.018 \\
\rowcolor{gray!20}
\cellcolor{white} & \textbf{Ctrl-CWM} & \textbf{0.010} & \textbf{0.012} & \textbf{0.012} & \textbf{0.150} & 0.866 & \underline{0.822} & \textbf{0.304} & \underline{0.506} \\
      \midrule
      \multirow{4}{*}{ZARA2} & ORCA & 0.013 & \textbf{0.007} & \textbf{0.007} & 0.136 & 1.017 & 2.501 & 0.101 & \textbf{0.025} \\
 & TRACE & 0.014 & \underline{0.011} & \underline{0.011} & 0.139 & \textbf{0.223} & 0.707 & \underline{0.337} & 1.502 \\
 & CrowdES & \textbf{0.009} & 0.013 & 0.013 & \textbf{0.100} & \underline{0.227} & \textbf{0.579} & \textbf{0.355} & 1.021 \\
\rowcolor{gray!20}
\cellcolor{white} & \textbf{Ctrl-CWM} & \underline{0.012} & \underline{0.011} & \underline{0.011} & \underline{0.124} & 0.505 & \underline{0.675} & 0.334 & \underline{1.000} \\
      \midrule
      \multirow{4}{*}{SDD} & ORCA & 0.052 & 0.036 & 0.032 & 0.421 & \underline{0.674} & \textbf{5.674} & 0.311 & 0.392 \\
 & TRACE & \underline{0.039} & \underline{0.030} & \underline{0.029} & \underline{0.404} & 0.712 & \underline{6.145} & \underline{0.328} & \underline{0.135} \\
 & CrowdES & \textbf{0.038} & 0.033 & 0.030 & 0.463 & \textbf{0.650} & 6.352 & \textbf{0.354} & 0.411 \\
\rowcolor{gray!20}
\cellcolor{white} & \textbf{Ctrl-CWM} & \textbf{0.038} & \textbf{0.028} & \textbf{0.028} & \textbf{0.373} & 0.907 & 6.165 & 0.319 & \textbf{0.123} \\
      \midrule
      \multirow{4}{*}{GCS} & ORCA & 6.677 & 0.439 & 0.439 & 15.429 & 2.097 & 4.558 & 0.240 & \textbf{0.004} \\
 & TRACE & 3.284 & 0.634 & 0.634 & 7.587 & \textbf{0.241} & \underline{1.875} & \underline{0.368} & 9.073 \\
 & CrowdES & \underline{0.584} & \underline{0.066} & \underline{0.066} & \textbf{0.329} & 1.341 & 3.864 & 0.279 & 8.519 \\
\rowcolor{gray!20}
\cellcolor{white} & \textbf{Ctrl-CWM} & \textbf{0.340} & \textbf{0.056} & \textbf{0.056} & \underline{0.870} & \underline{1.325} & \textbf{1.547} & \textbf{0.414} & \underline{6.080} \\
      \midrule
      \multirow{4}{*}{\textbf{AVG}} & ORCA & 1.029 & 0.106 & 0.106 & 2.595 & 1.198 & 3.228 & 0.180 & \textbf{0.065} \\
 & TRACE & 0.532 & 0.131 & 0.131 & 1.327 & \textbf{0.359} & \underline{1.878} & \underline{0.300} & 2.179 \\
 & CrowdES & \underline{0.147} & \underline{0.052} & \underline{0.051} & \textbf{0.315} & \underline{0.525} & 2.126 & 0.297 & 1.930 \\
\rowcolor{gray!20}
\cellcolor{white} & \textbf{Ctrl-CWM} & \textbf{0.110} & \textbf{0.045} & \textbf{0.045} & \underline{0.359} & 0.651 & \textbf{1.762} & \textbf{0.309} & \underline{1.426} \\
      \bottomrule
    \end{tabular}%
 }
\vspace{-3mm}
\end{table*}

\subsection{Planning and Run-time Control}
\label{sec:planning}
At run time, the planner combines critic scores with user-defined spatial costs to select pedestrian displacements. Users can change the spatial objective during simulation without re-training.

\vspace{2mm}

\noindent\textbf{Candidate evaluation.}
Users specify regions to avoid or targets to approach. For each pedestrian $i$ selected for planning, we use the cross-entropy method (CEM)~\cite{de2005tutorial} to search for its next displacement. We initialize a Gaussian distribution with the actor proposal $\bar a_t^i$ as its mean. At each iteration, we sample $K$ candidate displacements and simulate the crowd for $H$ steps under each candidate, following \cref{sec:formulation}. The active pedestrian set remains fixed within each rollout.

For each candidate $k$, the critic assigns a motion score $q_k$ to pedestrian $i$. We compute a spatial cost $c_k$ from the predicted positions of all active pedestrians over $H$ steps. We subtract the spatial cost from the motion score as follows:
\begin{equation}
J_k = q_k - \lambda_{\mathrm{user}}c_k,
\label{eq:planning_score}
\end{equation}
where $\lambda_{\mathrm{user}}\geq 0$ is a balance factor for the spatial cost.
CEM favors candidates with higher $J_k$. For avoidance, $c_k$ is defined as the fraction of imagined pedestrian positions inside the specified region. For attraction, $c_k$ is the mean distance of these positions to the target center, normalized by the scene-grid width (Appendix C). Within each rollout, we cache the scene encoder's feature maps computed from the scene map and initial pedestrian heatmaps, while updating interaction features at each step. This approximates the full feature recomputation used in training the actor.

\vspace{2mm}

\noindent\textbf{Action selection and run-time control.}
At each iteration, CEM updates the Gaussian mean and variance using the highest-scoring quarter of candidates ranked by $J_k$. After $I$ iterations, we use the final Gaussian mean as the planned displacement $a_t^i$. We plan each selected pedestrian's displacement independently from the same crowd state. In each rollout, the actor predicts the other pedestrians' displacements. We execute all planned displacements simultaneously using~\cref{eq:transition} and replan at the next simulation step.

\section{Experiments}
\label{sec:exp}

\noindent\textbf{Datasets and benchmarks.}
We evaluate crowd motion realism and run-time control on ETH--UCY~\cite{pellegrini2009you,lerner2007crowds}, SDD~\cite{robicquet2016learning}, and GCS~\cite{yi2015understanding,gcs2}, widely used public pedestrian trajectory benchmarks. We additionally evaluate robot-reactive crowd control in Social-HM3D~\cite{gong2025cognition}, built on Habitat 3.0~\cite{puig2024habitat}.

\vspace{2mm}

\noindent\textbf{Metrics.} 
To assess how closely generated crowds reproduce pedestrian motion observed in real-world videos, we adopt the scene-level and agent-level metrics of CrowdES~\cite{bae2025crowdes}. Scene-level metrics evaluate crowd distributions through density (Dens.), frequency (Freq.), coverage (Cov.), and population (Pop.). Agent-level metrics assess individual motion through kinematics (Kinem.), trajectory similarity using Dynamic Time Warping (DTW), diversity (Div.), and inter-agent collision rate (Col.). We use Earth Mover's Distance (EMD) to compare generated and observed distributions for all four scene-level metrics and Kinem. For run-time control, we define compliance as the normalized reduction in region occupancy for avoidance or increase for attraction, relative to matched uncontrolled runs. Detailed generation and control metrics are given in Appendices B.2 and C.1, respectively.

\subsection{Crowd Behavior Generation}
\label{sec:exp_generation}
We compare Ctrl-CWM with the rule-based baseline ORCA~\cite{van2010optimal} and the learning-based methods TRACE~\cite{rempe2023trace} and CrowdES~\cite{bae2025crowdes} under random-surface and diffusion arrivals. Emitters determine when and where pedestrians enter, with identical arrivals replayed across methods to isolate differences in motion generation and interactions. Ctrl-CWM uses critic-guided planning for all active pedestrians ($\lambda_{\mathrm{user}}=0$), with local goals from the pretrained goal head.

Table~\ref{tab:gen_shared} shows that Ctrl-CWM ranks first on nine of sixteen average scores, achieving the lowest average DTW, frequency error, and coverage error under both arrival protocols. With identical arrivals, these gains indicate that our model reproduces real-world pedestrian motion. ORCA achieves the lowest average collision rate but the highest density error and DTW, showing that collision avoidance alone does not ensure agreement with observed crowd motion.

Relative to CrowdES, Ctrl-CWM reduces average DTW by 22\% and 17\% under random-surface and diffusion arrivals, respectively, alongside collision-rate reductions of 50\% and 26\%. Thus, Ctrl-CWM generates more realistic crowd trajectories with fewer collisions. Ctrl-CWM also achieves lower DTW than TRACE on five of seven benchmark entries under random-surface arrivals and six under diffusion arrivals, demonstrating consistent trajectory improvements over TRACE across the two arrival protocols.

\begin{table}[t]
\centering
\caption{\textbf{Run-time avoidance control.}
Ctrl-CWM activates objectives online.
Baseline controls are active from initialization.
Higher compliance is better.
\textbf{Bold}/\underline{underline}: best/second-best compliance.}
\vspace{-2mm}
\label{tab:control}
\scriptsize
\setlength{\tabcolsep}{2pt}
\resizebox{\columnwidth}{!}{%
\begin{tabular}{lcccccccc}
\toprule
Method & ETH & HOTEL & UNIV & ZARA1 & ZARA2 & SDD & GCS & \textbf{AVG} \\
\midrule
\multicolumn{9}{@{}l}{\textit{Disc}} \\
ORCA
& \textbf{1.000} & $-$1.154 & 0.530 & 0.428
& \textbf{1.000} & 0.344 & \underline{0.999} & 0.450 \\
TRACE
& \underline{0.999} & \underline{0.898} & \underline{0.864}
& \underline{0.798} & \underline{0.965} & \underline{0.988}
& 0.996 & \underline{0.930} \\
CrowdES
& 0.599 & 0.546 & 0.727 & 0.469
& 0.598 & 0.317 & 0.436 & 0.527 \\
\rowcolor{gray!20}
\textbf{Ctrl-CWM}
& \textbf{1.000} & \textbf{0.998} & \textbf{0.986}
& \textbf{0.994} & \textbf{1.000} & \textbf{0.993}
& \textbf{1.000} & \textbf{0.996} \\
\midrule
\multicolumn{9}{@{}l}{\textit{Rectangle}} \\
ORCA
& \textbf{1.000} & $-$1.484 & 0.526 & $-$0.049
& \textbf{1.000} & $-$1.827 & 0.595 & $-$0.034 \\
TRACE
& \textbf{1.000} & \underline{0.896} & \underline{0.857}
& \underline{0.812} & 0.931 & \underline{0.976}
& \underline{0.988} & \underline{0.923} \\
CrowdES
& 0.883 & 0.808 & 0.779 & 0.790
& 0.824 & 0.114 & 0.301 & 0.643 \\
\rowcolor{gray!20}
\textbf{Ctrl-CWM}
& \textbf{1.000} & \textbf{0.996} & \textbf{0.939}
& \textbf{0.996} & \underline{0.999} & \textbf{0.996}
& \textbf{1.000} & \textbf{0.989} \\
\midrule
\multicolumn{9}{@{}l}{\textit{Multiple}} \\
ORCA
& \underline{0.973} & \underline{0.999} & 0.601 & $-$1.952
& $-$0.304 & $-$0.473 & 0.768 & 0.087 \\
TRACE
& \textbf{1.000} & 0.932 & \underline{0.937}
& \underline{0.867} & \underline{0.827} & \textbf{0.986}
& \underline{0.983} & \underline{0.933} \\
CrowdES
& 0.479 & 0.542 & 0.549 & 0.781
& 0.751 & 0.525 & 0.424 & 0.579 \\
\rowcolor{gray!20}
\textbf{Ctrl-CWM}
& \textbf{1.000} & \textbf{1.000} & \textbf{0.996}
& \textbf{0.989} & \textbf{0.994} & \underline{0.983}
& \textbf{1.000} & \textbf{0.995} \\
\bottomrule
\end{tabular}%
}
\vspace{-2mm}
\end{table}

\begin{table}[t]
\centering
\caption{\textbf{Runtime per simulated second.}
Measured under avoidance control, with actor-only runtime in parentheses.}
\label{tab:latency}
\vspace{-2mm}
\small
\renewcommand{\arraystretch}{1.1}
\setlength{\tabcolsep}{4pt}
\resizebox{\columnwidth}{!}{%
\begin{tabular}{lcccc}
\toprule
Method & ORCA & TRACE & CrowdES
& \cellcolor{gray!20}\textbf{Ctrl-CWM} \\
\midrule
Time (ms/simulated s)$\downarrow$
& 71 & 1{,}213 & 87
& \cellcolor{gray!20}104\,(32) \\
\bottomrule
\end{tabular}%
}
\vspace{-4mm}
\end{table}

\subsection{Run-time Spatial Control}
\label{sec:exp_control}
\paragraph{Protocol.} Ctrl-CWM supports user-defined spatial objectives for
static regions and targets, as well as regions attached to moving entities such as robots. We evaluate avoidance using disc-shaped, rectangular, and multiple disjoint zones along each scene's main pedestrian flow
(see Appendix C). ORCA treats the zones as obstacles, TRACE applies test-time avoidance guidance, and CrowdES uses map constraints. Ctrl-CWM incorporates objectives through the user cost in Eq.~\ref{eq:planning_score}. Controlled and uncontrolled Ctrl-CWM runs share the first third of each episode before avoidance is activated only in the controlled run. Baseline controls are active from initialization. All methods use identical arrivals, with 10 trials per scene evaluated over each episode's final two thirds.


\begin{figure}[t]
    \centering
    \includegraphics[width=\columnwidth]{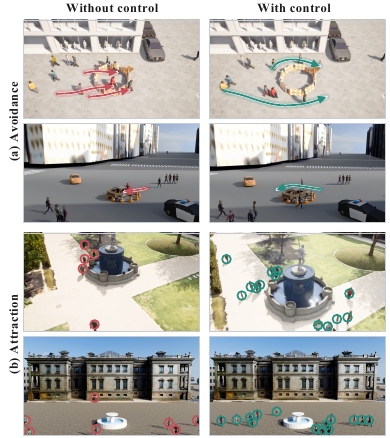}
    \vspace{-6mm}
    \caption{\textbf{Run-time control.}
    The left and right pairs show uncontrolled (red) and controlled (green) rollouts, respectively.
    (a) An avoidance objective redirects pedestrians away from the hazardous zone.
    (b) An attraction objective guides them toward the fountain.}
    \label{fig:runtime_control}
    \vspace{-5mm}
\end{figure}

\noindent\textbf{Quantitative results.}
Table~\ref{tab:control} reports avoidance compliance, measured as the relative reduction in the fraction of pedestrian steps inside avoidance regions compared with the corresponding uncontrolled run. A score of 1 indicates zero controlled occupancy, 0 indicates no reduction, and negative values indicate increased occupancy. Ctrl-CWM achieves the highest average compliance across all three zone configurations (0.989--0.996) and ranks first, including ties, in 19 of 21 scene--configuration comparisons.

To compare computational efficiency, we measure runtime in milliseconds per simulated second of crowd motion. As shown in Table~\ref{tab:latency}, ORCA and CrowdES are faster than the full Ctrl-CWM model, taking 71 and 87\,ms, respectively. In this evaluation, ORCA and CrowdES use avoidance constraints specified at initialization, which remain fixed during simulation. Ctrl-CWM supports objectives introduced during simulation and takes 104\,ms with planning and 32\,ms for actor-only generation. It also achieves higher average compliance than TRACE across all three zone configurations at substantially lower runtime (104\,ms vs. 1213\,ms).

\noindent\textbf{Qualitative results.}
We visualize the generated pedestrian trajectories in CARLA~\cite{dosovitskiy2017carla} using Unreal Engine~\cite{unrealsoftware}. Figure~\ref{fig:runtime_control} shows pedestrians redirected away from hazardous zones under avoidance and toward a fountain under attraction. Quantitative attraction results and the compliance--realism trade-off are provided in Appendix C.2. Humanoid animations serve only to visualize trajectory-level control; character animation quality is outside the scope of this work.


\subsection{Robot-Reactive Crowd Control in Habitat 3.0}
\label{sec:habitat}
We evaluate whether Ctrl-CWM can control pedestrian responses to a moving robot and how these responses affect navigation performance.

\vspace{2mm}

\noindent\textbf{Setup.}
We use Social-HM3D~\cite{gong2025cognition}, built on Habitat 3.0~\cite{puig2024habitat}, where a Spot robot navigates to a point goal among waypoint-following humanoids. We evaluate the A$^{*}$~\cite{hart1968formal} and ORCA planners~\cite{van2011reciprocal} and the released Falcon policy~\cite{gong2025cognition} on 1{,}087 validation episodes across 100 scenes, keeping each navigation method fixed across crowd conditions. Using identical pedestrian waypoints and episode settings, we compare benchmark humanoids (Original) with Ctrl-CWM under attraction, no user objective (None), and avoidance.

We apply the GCS-trained model to these unseen indoor scenes without retraining or fine-tuning on Habitat data. Ctrl-CWM periodically replans nearby pedestrians using a robot-centered zone for attraction or avoidance. We blend the planned motion with waypoint-following motion for execution by the humanoid controller. Pedestrians outside the replanning radius follow their nominal routes. Integration details are provided in Appendix D.


\begin{table}[t]
\centering
\caption{\textbf{Run-time crowd control in Social-HM3D.}
Pink/blue indicates lower/higher SR, SPL, and PSC
and higher/lower H-coll. and Zone relative to None.
All values except Steps are percentages.}
\vspace{-2mm}
\label{tab:nav_habitat}
\small
\setlength{\tabcolsep}{3pt}
\resizebox{\columnwidth}{!}{%
\begin{tabular}{@{}lllcccccc@{}}
\toprule
Method & Crowd & Mode
& SR$\uparrow$ & SPL$\uparrow$ & PSC$\uparrow$
& H-coll.$\downarrow$ & Zone & Steps \\
\midrule
\multirow{4}{*}{A$^{*}$~\cite{hart1968formal}}
 & Original & --
 & 45.6 & 44.8 & 90.4 & 54.2 & -- & 73.6 \\
 & \multirow{3}{*}{Ctrl-CWM} & Attract
 & \pinkcell{36.8} & \pinkcell{36.1} & \pinkcell{87.3}
 & \pinkcell{62.9} & \pinkcell{24.7} & 65.6 \\
 & & None
 & 47.0 & 46.1 & 89.5 & 52.7 & 8.9 & 72.1 \\
 & & Avoid
 & \bluecell{60.7} & \bluecell{59.6} & \bluecell{91.9}
 & \bluecell{39.5} & \bluecell{0.4} & 80.5 \\
\midrule
\multirow{4}{*}{ORCA~\cite{van2011reciprocal}}
 & Original & --
 & 35.7 & 31.1 & 92.9 & 37.2 & -- & 226.1 \\
 & \multirow{3}{*}{Ctrl-CWM} & Attract
 & \pinkcell{9.8} & \pinkcell{9.2} & \pinkcell{83.3}
 & \pinkcell{59.4} & \pinkcell{36.3} & 267.0 \\
 & & None
 & 24.6 & 22.3 & 93.9 & 23.1 & 6.4 & 325.3 \\
 & & Avoid
 & \bluecell{48.9} & \bluecell{43.7} & \bluecell{97.2}
 & \bluecell{12.7} & \bluecell{0.3} & 303.8 \\
\midrule
\multirow{4}{*}{Falcon~\cite{gong2025cognition}}
 & Original & --
 & 54.4 & 49.8 & 89.7 & 42.5 & -- & 107.5 \\
 & \multirow{3}{*}{Ctrl-CWM} & Attract
 & \pinkcell{45.2} & \pinkcell{42.0} & \pinkcell{84.6}
 & \pinkcell{54.4} & \pinkcell{26.4} & 90.2 \\
 & & None
 & 59.9 & 55.1 & 88.2 & 38.0 & 8.8 & 105.7 \\
 & & Avoid
 & \bluecell{75.5} & \bluecell{69.0} & \bluecell{91.9}
 & \bluecell{20.8} & \bluecell{0.3} & 122.0 \\
\bottomrule
\end{tabular}%
}
\vspace{-4mm}
\end{table}

\vspace{2mm}
\noindent\textbf{Metrics.}
Zone measures the fraction of replanned pedestrian steps ending inside the robot-centered zone. Relative to None, attraction should increase Zone and avoidance decrease it. To assess downstream task performance, we report social navigation metrics~\cite{gong2025cognition}: success rate (SR), success weighted by path length (SPL), personal-space compliance (PSC), the fraction of episodes with a human collision (H-coll.), and mean episode length in steps (Steps).

\noindent\textbf{Quantitative results.}
Table~\ref{tab:nav_habitat} shows that attraction and avoidance induce contrasting pedestrian responses to a moving robot across all three navigation methods. Relative to None, attraction increases robot-zone occupancy from 6.4--8.9\% to 24.7--36.3\%, while avoidance reduces it to 0.3--0.4\%. These contrasting responses lead to robot--pedestrian interactions: attraction draws pedestrians toward the robot, potentially obstructing its path, whereas avoidance encourages them to move aside and yield passage. Accordingly, attraction degrades navigation performance, while avoidance improves success, path efficiency, and personal-space compliance and reduces human collisions compared with both None and Original. These results demonstrate that Ctrl-CWM can shape robot--pedestrian interactions through user-defined attraction and avoidance, creating diverse navigation conditions without modifying the robot policies.

\begin{figure}[t]
    \centering
    \includegraphics[width=\columnwidth]{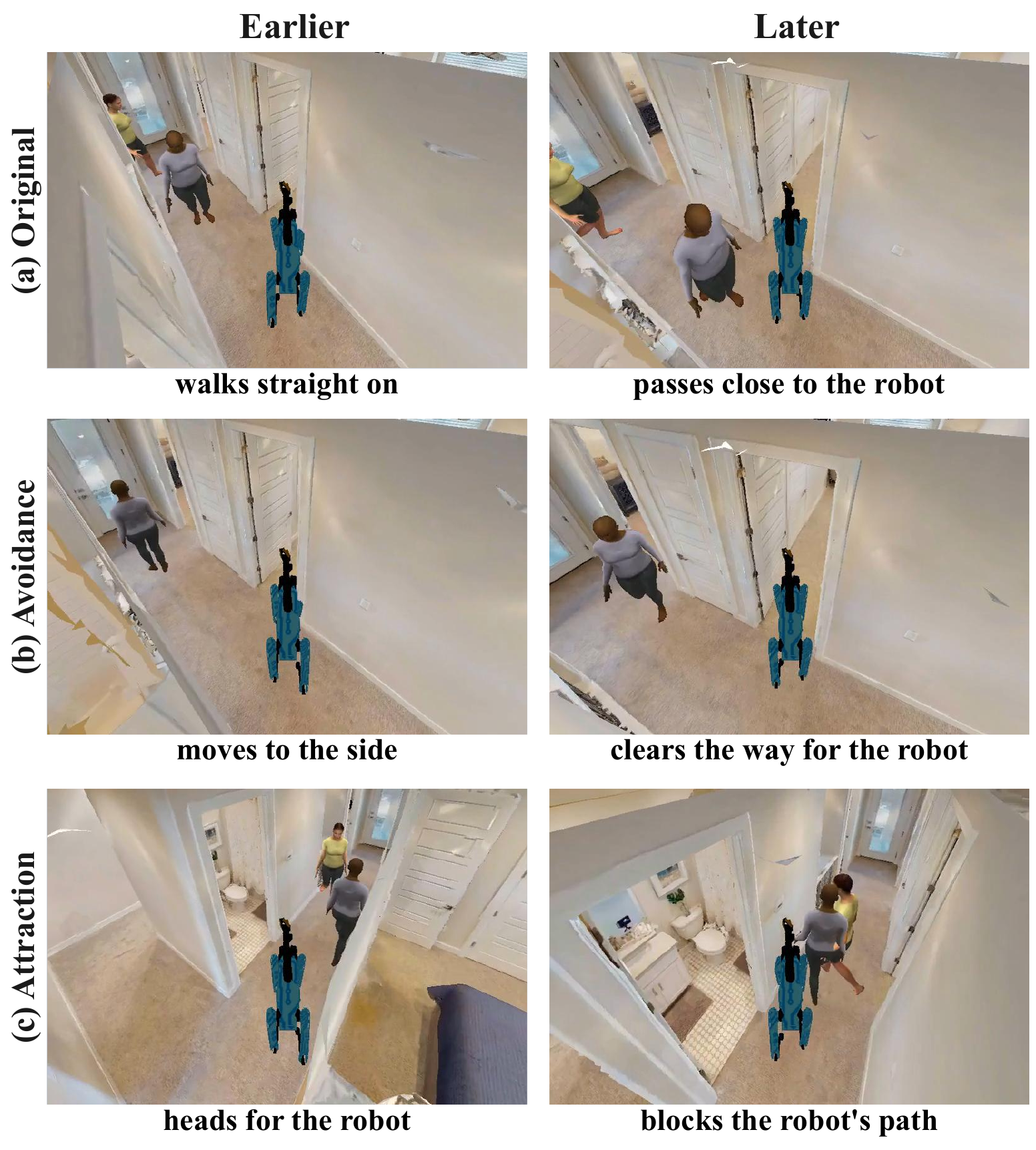}
    \vspace{-7mm}
    \caption{\textbf{Qualitative crowd control in Habitat 3.0.}
Columns show earlier and later states.
Top: an original humanoid approaches and passes close to the robot.
Middle: under avoidance, the pedestrian moves aside, clearing the robot's path.
Bottom: under attraction, the pedestrian approaches the robot and blocks its path.}
    \label{fig:habitat_control}
    \vspace{-4mm}
\end{figure}

\begin{figure*}[!t]
  \centering
  \includegraphics[width=\textwidth]{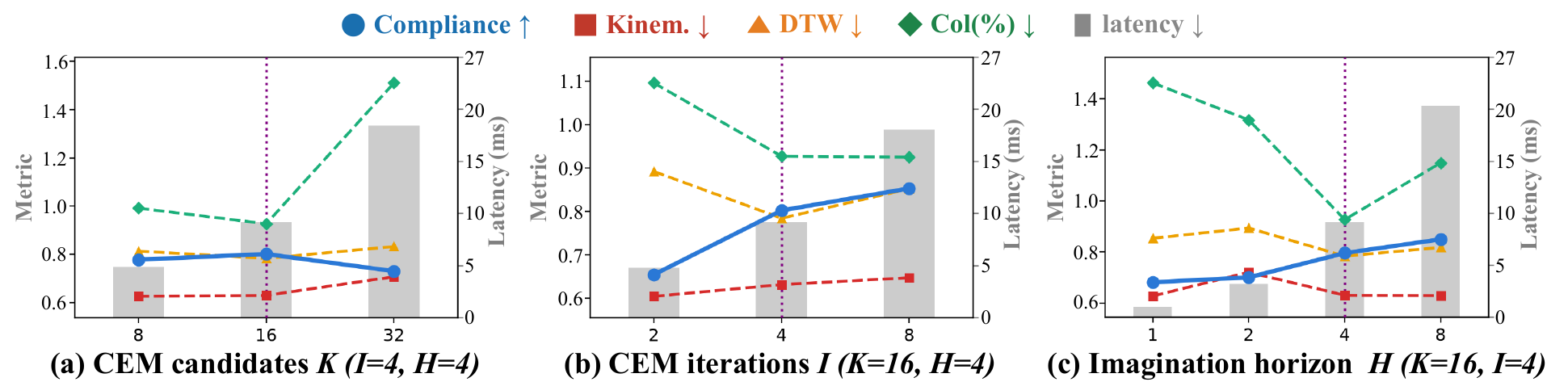}
  \vspace{-6mm}
\caption{\textbf{Ablation of planning parameters.}
Lines show compliance and crowd-motion metrics over
the post-activation control window. Gray bars report
per-call latency in milliseconds on the right axes.
Vertical dotted lines mark default settings.}
  \label{fig:planning_ablation}
  \vspace{-5mm}
\end{figure*}

\vspace{2mm}

\noindent\textbf{Qualitative results.}
Figure~\ref{fig:habitat_control} illustrates different robot--pedestrian interactions in an episode. With the original humanoids, a pedestrian approaches and passes close to the robot. Under avoidance, the pedestrian moves aside, clearing the robot's path. Under attraction, the pedestrian instead approaches the robot and blocks its path, preventing it from reaching the goal. This example illustrates how Ctrl-CWM can generate diverse pedestrian responses to a moving robot by changing only the crowd behavior. These results highlight its potential as a controllable crowd simulator for creating diverse robot--pedestrian interaction scenarios and evaluating robot navigation under different conditions.

\subsection{User Study}

\label{sec:user_study}
We conducted a blinded study with 100 participants from CloudResearch Connect~\cite{hartman2023introducing} to assess perceived crowd controllability. In Part 1, participants compared Ctrl-CWM and CrowdES in four scenes, selecting which crowd better followed the objective. In Part 2, participants independently labeled clips from eight Habitat episodes across four scenes as yielding to, obstructing, or ignoring the robot. These Ctrl-CWM clips used avoidance, attraction, or no user objective.

Ctrl-CWM was selected in 85.8\% of comparisons with CrowdES (Table~\ref{tab:user_study}). Behavior-label agreement in Habitat was 78.8--85.4\% across the three conditions. These results show that users recognize objective-directed crowd behavior and intended pedestrian responses to moving robots.

\begin{table}[t]
\centering
\caption{\textbf{User evaluation of crowd controllability.}
Objective following reports selection rates between
Ctrl-CWM and CrowdES. Behavior identification reports label agreement for
Ctrl-CWM under avoidance, attraction, and no user objective (None).}
\label{tab:user_study}
\vspace{-3mm}
\small
\setlength{\tabcolsep}{3pt}
\begin{tabular*}{\columnwidth}
{@{\extracolsep{\fill}}ccccc@{}}
\toprule
\multicolumn{2}{c}{Part 1. Objective following}
& \multicolumn{3}{c}{Part 2. Behavior identification} \\
\cmidrule(lr){1-2}\cmidrule(lr){3-5}
Ctrl-CWM & \shortstack{CrowdES}
& Avoid & Attract & None \\
\midrule
85.8\% & 14.3\% & 85.4\% & 78.8\% & 81.8\% \\
\bottomrule
\end{tabular*}
\vspace{-5mm}
\end{table}

\subsection{Ablation Studies}
\label{sec:ablation}
We examine prediction pretraining, critic-guided planning, and training objectives in Table~\ref{tab:ablation}, and planning parameters in Fig.~\ref{fig:planning_ablation}. Additional analyses are provided in Appendix F.

\noindent\textbf{Prediction pretraining.}
We replace the pretrained crowd representation module with a randomly initialized counterpart while retaining the same behavior-learning procedure. Pretraining reduces kinematic error from 1.205 to 0.465 and increases diversity from 0.157 to 0.285. These gains indicate that features learned through trajectory prediction benefit subsequent crowd generation. Trajectory-prediction results are provided in Appendix G.

\noindent\textbf{Planning and learned ranking.}
Critic-guided planning reduces DTW from 0.970 to 0.925 and collision rate from 0.931\% to 0.756\% relative to actor-only generation, with kinematic error increasing from 0.432 to 0.465. Under the same control protocol, ranking candidates by user cost alone achieves higher compliance than the full model (0.858 vs.\ 0.824) but performs worse on all generation metrics. These results highlight the critic's role in balancing avoidance compliance and generation quality.

\begin{table}[t]
\centering
\caption{\textbf{Component ablations under diffusion arrivals.}
Results are averaged over five ETH--UCY folds.
Freq.\ and Cov.\ coincide and are reported once.
Comp.: avoidance compliance. N/A: control not evaluated.
Col.\ is in \%. \textbf{Bold}/\underline{underline}: best/second-best.}
\label{tab:ablation}
\vspace{-3mm}
\scriptsize
\setlength{\tabcolsep}{2pt}
\begin{tabular}{@{}l*{8}{c}@{}}
    \toprule
    Variant
    & \multicolumn{3}{c}{Scene-level}
    & \multicolumn{4}{c}{Agent-level}
    & Control \\
    \cmidrule(lr){2-4}\cmidrule(lr){5-8}\cmidrule(l){9-9}
    & Dens.$\downarrow$
    & \makecell{Freq./\\Cov.$\downarrow$}
    & Pop.$\downarrow$
    & Kinem.$\downarrow$
    & DTW$\downarrow$
    & Div.$\uparrow$
    & Col.$\downarrow$
    & Comp.$\uparrow$ \\
    \midrule
    Random encoder
    & 0.101 & 0.067 & 0.304
    & 1.205 & 1.031 & 0.157 & 1.101
    & \textcolor{gray}{N/A} \\
    \midrule
    Actor only
    & \textbf{0.077} & \textbf{0.047} & 0.271
    & \textbf{0.432} & 0.970 & 0.280 & 0.931
    & \textcolor{gray}{N/A} \\
    User cost only
    & 0.089 & 0.075 & 0.294
    & 0.812 & 1.178 & 0.078 & 2.398
    & \textbf{0.858} \\
    \rowcolor{gray!12}
    \textbf{Full model}
    & \underline{0.078} & \textbf{0.047} & \textbf{0.254}
    & \underline{0.465} & \textbf{0.925} & \textbf{0.285} & \textbf{0.756}
    & \underline{0.824} \\
    \addlinespace[2pt]
    \;w/o soft-DTW
    & 0.082 & 0.051 & 0.275
    & 0.846 & 0.989 & 0.182 & 0.838
    & \textcolor{gray}{N/A} \\
    \;w/o collision
    & 0.088 & \underline{0.049} & \underline{0.266}
    & 0.472 & 0.965 & \underline{0.283} & 0.919
    & 0.645 \\
    \;w/o kinematic
    & 0.081 & 0.053 & 0.271
    & 0.502 & \underline{0.933} & 0.281 & \underline{0.817}
    & 0.772 \\
    \bottomrule
\end{tabular}
\vspace{-4mm}
\end{table}

\noindent\textbf{Training objectives.}
We examine how individual losses affect motion quality. Removing soft-DTW increases DTW to 0.989 and kinematic error to 0.846 while reducing diversity to 0.182. Removing collision supervision reduces compliance to 0.645 and raises collisions to 0.919\%, while removing kinematic supervision reduces compliance to 0.772 and increases kinematic error to 0.502. These results support complementary roles in trajectory matching, collision avoidance, and motion dynamics.

\noindent\textbf{Effect of planning parameters.}
Figure~\ref{fig:planning_ablation} varies candidate count $K$, CEM iterations $I$, and planning horizon $H$. Among candidate counts, $K=16$ achieves the highest compliance and lowest DTW and collision rate. Increasing $I$ or $H$ from 4 to 8 raises compliance and latency, but also increases DTW. At $H=8$, collisions increase relative to $H=4$. At $H=1$, latency decreases, but compliance drops and collisions increase. Thus, the default $K=16$, $I=4$, $H=4$ balances these criteria at about 10\,ms per call.
\section{Conclusion}
We introduce Ctrl-CWM for crowd trajectory generation and run-time control without retraining. Experiments show improved motion quality, evaluated against ground-truth trajectories, and effective avoidance and attraction, including controllable pedestrian responses to navigating robots.

\noindent\textbf{Limitations.}
We expect to improve the practical utility of Ctrl-CWM, which will be the first step toward building a world model to simulate social phenomena in the real world. To do this, we need to represent and understand more complex social behaviors and interactions between agents, such as handshakes and group conversations.

\section*{LLM Usage Statement}
We used ChatGPT and Claude for grammatical editing and assistance with implementing baseline methods. These tools were not used to formulate the research idea, conduct experiments or comparisons, or analyze the results. We reviewed all text and code produced with their assistance and take full responsibility for the final content.

{
    \small
    \bibliographystyle{ieeenat_fullname}
    \bibliography{main}
}

\begin{bibunit}[ieeenat_fullname]
    \backrefsetup{disable}
    \maketitlesupplementary
\appendix
\makeatletter
\setlength{\@fptop}{0pt}
\makeatother

This supplementary material provides additional implementation details and experimental results. Appendix~\ref{app:implementation} details the model architecture, training objectives, and planning procedures. Appendix~\ref{app:protocols} describes the evaluation protocols and generation metrics. Appendix~\ref{app:control} defines the spatial costs and compliance metrics and examines how control varies with the cost weight and crowd population. Appendices~\ref{app:habitat} and \ref{app:user_study} describe the Habitat simulation and user study. We also provide ablations, trajectory prediction results, and qualitative examples in Appendices~\ref{app:ablation}--\ref{app:qualitative}.

\section{Implementation Details}
\label{app:implementation}
We follow the notation in Secs.~\ref{sec:formulation}--\ref{sec:planning}: $s_t$ denotes the crowd state, $z_t^i$ the crowd representation, $\bar a_t^i$ the displacement proposed by the actor, and $a_t^i$ the displacement selected by the planner. We additionally denote pedestrian $i$'s local goal by $g_t^i$, its velocity history by $\mathcal H_t^i$, and the active pedestrian set by $\mathcal A_t$. A hat denotes a simulated quantity, a star denotes ground truth, and $(k)$ identifies a candidate rollout. For a fixed state and pedestrian, $y_k$, $q_k$, $c_k$, and $J_k$ denote the training cost, critic score, spatial cost, and planning score, respectively.

\subsection{Model Architecture}
\label{app:arch}
\paragraph{Scene representation.}
We map the scene geometry and pedestrian positions to a $G\times G$ grid using a scene-specific similarity transform, with $G=192$. Positions, displacements, and distances are expressed in grid pixels, except where stated otherwise. The static scene map contains a binary walkability mask and eight semantic channels: background, building, structure, bush, grass, tree, sidewalk, and road. These follow the CrowdES categories with an additional background channel~\cite{bae2025crowdes}.
Two dynamic channels contain a heatmap for pedestrian $i$ and a summed heatmap of the pedestrians currently present in the scene. Both use Gaussians with standard deviation 5 grid pixels and are aligned with the scene map, giving eleven input channels in total.

\paragraph{Scene and interaction encoders.}
The U-Net uses width multiplier $w=0.25$, with skip features of 8, 16, 32, and 64 channels.
Four pooling stages produce a $64\times12\times12$ bottleneck.
We sample an eight-dimensional local scene feature $\ell_t^i$ at $p_t^i$
from the full-resolution encoder feature map.
The interaction encoder embeds each pedestrian's 2D position into a 64-dimensional vector. A separate $2\to16\to1$ MLP computes an attention bias for each pedestrian pair from their relative position.
A $2\to64$ GRU encodes the velocity history $\mathcal H_t^i$.
The attention output and history state are concatenated and projected through
$128\to96\to96$ layers to obtain an interaction feature $c_t^i\in\mathbb R^{96}$.
A FiLM layer maps this feature to channel-wise scales and shifts for the scene bottleneck
used by the prediction heads.
The actor-facing crowd representation is
$z_t^i=[\ell_t^i\|c_t^i]\in\mathbb R^{104}$.
Thus, $\phi_{\mathrm{rep}}$ supplies scene context, recent motion, and pedestrian interactions
through the interface defined in Eq.~\ref{eq:components}.
Table~\ref{tab:network_structure_wm} details the architecture.

\paragraph{Prediction heads.}
The goal head decodes the modulated bottleneck and scene skip features into a heatmap for the endpoint of the $L$-step prediction window. This endpoint serves as a local goal and need not be the pedestrian's final destination. The waypoint and auxiliary heads share a goal-conditioned decoder trunk and have separate $1\times1$ output layers. A four-level goal encoder produces 16, 32, 64, and 128 channels. At each level of the shared decoder trunk, upsampled features are concatenated with scene and goal features, then processed by a $3\times3$ convolution, GroupNorm with eight groups, and SiLU. The waypoint head outputs $L=12$ heatmaps. The auxiliary head outputs one next-position heatmap. Both heads are conditioned on the ground-truth endpoint during pretraining. During forecasting, the waypoint head uses a predicted endpoint. The auxiliary head provides pretraining supervision. Actor-generated displacements advance the crowd during subsequent simulation.

\paragraph{Actor.}
The shared actor receives the crowd representation, local-goal direction, and current velocity:
\begin{equation*}
[z_t^i\|u_t^i\|\tilde v_t^i]\in\mathbb R^{108},
\end{equation*}
where $\tilde v_t^i=v_t^i/G$ is normalized internally before concatenation.
For $g_t^i\ne p_t^i$, the direction is
$u_t^i=(g_t^i-p_t^i)/\|g_t^i-p_t^i\|_2$.
The actor maps this input to the next-step displacement $\bar a_t^i$ using two 128-unit hidden layers with ReLU activations and a two-dimensional output layer.

\paragraph{Critic and discriminator.}
For a pedestrian in a crowd state, we construct six geometric features:
\begin{equation*}
\begin{aligned}
\bigl[\,&v_x/G,\ v_y/G,\ d_{\mathrm{nbr}}/G,\ d_{\mathrm{goal}}/G,\\
&\cos(v,g-p),\ \|v\|_2/G\,\bigr].
\end{aligned}
\end{equation*}
Here, $p$, $v$, and $g$ denote that pedestrian's position, velocity, and local goal. The distances to the nearest neighbor and local goal are denoted by $d_{\mathrm{nbr}}$ and $d_{\mathrm{goal}}$, respectively. The cosine measures alignment between its motion and goal direction. For candidate $k$, we compute these features for pedestrian $i$ in each state of $\hat s_{t+1:t+H}^{(k)}$. A $6\to128$ GRU encodes the resulting $H$-step sequence, and a $128\to128\to1$ MLP maps the final hidden state to $q_k$. The critic has approximately 69K parameters and uses rollout geometry without directly receiving $z_t^i$.

The training discriminator uses a $2\to128$ GRU and a $128\to128\to1$ MLP with LeakyReLU(0.2). It scores sequences of $L$ normalized displacement vectors, including the displacement from the observed position to the first prediction. Table~\ref{tab:network_structure_behavior} summarizes these networks.

\begin{table*}[t]
\caption{\textbf{Crowd representation and prediction-head interfaces.}
Spatial dimensions are per pedestrian, with batch axes omitted.
$A$ is the number of co-present agents and $T_h$ is the velocity-history length.
Widths use $w=0.25$. GN(8) denotes GroupNorm with eight groups.}
\label{tab:network_structure_wm}
\centering\footnotesize
\setlength{\tabcolsep}{3pt}
\renewcommand{\arraystretch}{1.08}
\begin{tabular}{@{}p{.06\linewidth}p{.39\linewidth}p{.24\linewidth}p{.25\linewidth}@{}}
\toprule
ID & Operation & Input & Output \\
\midrule
\multicolumn{4}{@{}l}{\textit{Inputs and scene encoder}}\\
1-1 & Similarity transform & Scene, trajectories & Grid, $G=192$ \\
1-2 & Walkability and semantic channels & Transformed scene & $9\times192\times192$ \\
1-3 & Individual Gaussian, $\sigma=5$ & Agent position & $1\times192\times192$ \\
1-4 & Crowd Gaussian accumulation & Crowd positions & $1\times192\times192$ \\
1-5 & Channel concatenation & 1-2, 1-3, 1-4 & $11\times192\times192$ \\
2-1 & Encoder block, 8 channels & 1-5 & $8\times192\times192$ \\
2-2 & Pool $2\times$, encoder block & 2-1 & $16\times96\times96$ \\
2-3 & Pool $2\times$, encoder block & 2-2 & $32\times48\times48$ \\
2-4 & Pool $2\times$, encoder block & 2-3 & $64\times24\times24$ \\
2-5 & Pool $2\times$, bottleneck & 2-4 & $64\times12\times12$ \\
2-6 & Local sampling at $p_t^i$ & 2-1, position & $\ell_t^i\in\mathbb R^8$ \\
\midrule
\multicolumn{4}{@{}l}{\textit{Interaction encoder}}\\
3-1 & Position embedding, linear $2\to64$ & Positions & $A\times64$ \\
3-2 & Pairwise bias, MLP $2\to16\to1$ & Relative positions & $A\times A\times1$ \\
3-3 & Attention with pairwise biases & 3-1, 3-2 & $A\times64$ \\
3-4 & History GRU $2\to64$ & $A\times T_h\times2$ & $A\times64$ \\
3-5 & Concatenation & 3-3, 3-4 & $A\times128$ \\
3-6 & Projection $128\to96\to96$, select pedestrian $i$ & 3-5 & $c_t^i\in\mathbb R^{96}$ \\
3-7 & FiLM scale and shift & 3-6 & Two 64-vectors \\
3-8 & Bottleneck modulation & 2-5, 3-7 & $64\times12\times12$ \\
\midrule
\multicolumn{4}{@{}l}{\textit{Goal features and prediction heads}}\\
4-1 & Goal decoder & 3-8, scene skips & $M_{\mathrm{goal}}$ \\
4-2 & Goal feature level 1 & Goal heatmap & $16\times192\times192$ \\
4-3 & Goal feature level 2 & 4-2 & $32\times96\times96$ \\
4-4 & Goal feature level 3 & 4-3 & $64\times48\times48$ \\
4-5 & Goal feature level 4 & 4-4 & $128\times24\times24$ \\
4-6 & Upsample, concatenate, conv, GN(8), SiLU & 3-8, 2-4, 4-5 & $64\times24\times24$ \\
4-7 & Upsample, concatenate, conv, GN(8), SiLU & 4-6, 2-3, 4-4 & $32\times48\times48$ \\
4-8 & Upsample, concatenate, conv, GN(8), SiLU & 4-7, 2-2, 4-3 & $16\times96\times96$ \\
4-9 & Upsample, concatenate, conv, GN(8), SiLU & 4-8, 2-1, 4-2 & $8\times192\times192$ \\
4-10 & Waypoint output, conv $1\times1$ & 4-9 & $L\times192\times192$ \\
4-11 & Auxiliary output, conv $1\times1$ & 4-9 & $1\times192\times192$ \\
\bottomrule
\end{tabular}
\end{table*}

\begin{table}[!htbp]
\caption{\textbf{Behavior-learning networks.}
The actor receives features from the frozen crowd representation module.
The critic uses $H$ imagined steps, while its training targets use $L$ steps.
The discriminator uses $L$ displacement vectors.}
\label{tab:network_structure_behavior}
\centering\small
\setlength{\tabcolsep}{4pt}
\renewcommand{\arraystretch}{1.12}
\begin{tabular}{@{}p{.23\linewidth}p{.43\linewidth}p{.22\linewidth}@{}}
\toprule
Module & Layers or input & Output \\
\midrule
Actor input & $[\ell_t^i\|c_t^i\|u_t^i\|\tilde v_t^i]$ & 108 features \\
Actor hidden & Linear $108\to128\to128$, ReLU & 128 features \\
Actor output & Linear $128\to2$ & Displacement \\
\midrule
Critic input & Six geometric features at each imagined step & $H\times6$ \\
Critic recurrent & GRU $6\to128$, final hidden state & 128 features \\
Critic output & MLP $128\to128\to1$ & Rollout score \\
\midrule
Discriminator input & Normalized displacement sequence & $L\times2$ \\
Discriminator recurrent & GRU $2\to128$, final hidden state & 128 features \\
Discriminator output & MLP $128\to128\to1$, LeakyReLU(0.2) & Sequence score \\
\bottomrule
\end{tabular}
\end{table}

\subsection{Training Procedure}
\label{app:training}
\paragraph{Prediction targets and losses.}
For pedestrian $i$ at time $t$, let $p_{t+j}^{i,*}$ be its ground-truth position $j$ steps ahead, for $j=1,\ldots,L$. This is the explicitly indexed form of $p_j^*$ in Sec.~\ref{sec:worldmodel}. Let $\mathcal G(p)$ denote a Gaussian target heatmap centered at $p$. The goal, waypoint, and auxiliary targets are
\begin{equation}
\begin{aligned}
M_{\mathrm{goal}}^\ast &= \mathcal G(p_{t+L}^{i,*}),\\
M_{\mathrm{way},j}^\ast &= \mathcal G(p_{t+j}^{i,*}),\\
M_{\mathrm{aux}}^\ast &= \mathcal G(p_{t+1}^{i,*}).
\end{aligned}
\label{eq:app_prediction_targets}
\end{equation}
The corresponding predicted heatmaps are denoted by
$M_{\mathrm{goal}}$, $M_{\mathrm{way},j}$, and $M_{\mathrm{aux}}$.
Let $\operatorname{WBCE}$ denote binary cross-entropy with positive-class weight 50.
The prediction losses are
\begin{equation}
\begin{aligned}
\ell(M,p^\ast)
&= \operatorname{WBCE}\bigl(M,\mathcal G(p^\ast)\bigr)\\
&\quad+\bigl\|\operatorname{softargmax}(M)-p^\ast\bigr\|_2,\\
\mathcal L_{\mathrm{goal}} &= \ell(M_{\mathrm{goal}},p_{t+L}^{i,*}),\\
\mathcal L_{\mathrm{way}} &= \frac{1}{L}\sum_{j=1}^{L}
\ell(M_{\mathrm{way},j},p_{t+j}^{i,*}),\\
\mathcal L_{\mathrm{aux}} &= \operatorname{WBCE}
(M_{\mathrm{aux}},M_{\mathrm{aux}}^\ast),\\
\mathcal L_{\mathrm{crowd}}
&= \mathcal L_{\mathrm{goal}}
+\lambda_{\mathrm{way}}\mathcal L_{\mathrm{way}}
+\lambda_{\mathrm{aux}}\mathcal L_{\mathrm{aux}}.
\end{aligned}
\label{eq:app_stage1}
\end{equation}
We use $\lambda_{\mathrm{way}}=1$ and $\lambda_{\mathrm{aux}}=0.5$. The goal and waypoint heads receive heatmap and coordinate supervision. The auxiliary head receives next-position heatmap supervision. The waypoint and auxiliary heads are conditioned on the ground-truth endpoint during pretraining.

\paragraph{Prediction pretraining.}
We jointly train $\phi_{\mathrm{rep}}$ and the prediction heads for 80 epochs using AdamW, a learning rate of $3\times10^{-4}$, weight decay $10^{-5}$, cosine annealing, and gradient-norm clipping at 1.0. We train on a single RTX 3090 with a batch size of one and gradient accumulation over three steps per optimizer update. After pretraining, $\phi_{\mathrm{rep}}$ remains frozen throughout subsequent actor and critic training.

\paragraph{Actor rollouts.}
We initialize each $L$-step rollout from an observed crowd state $s_t$ and assign each pedestrian a nearby future position sampled from its ground-truth trajectory as a training goal. At step $h=0,\ldots,L-1$, the actor predicts $\bar a_{t+h}^i$ for every pedestrian. We apply these displacements simultaneously, update positions and velocity histories, and recompute scene and interaction features from the resulting crowd state. The next prediction therefore uses the actor's generated state rather than a future ground-truth state.

\paragraph{Trajectory imitation.}
For pedestrian $i$ with observed position $p_t^i$, let $\tau=(\hat p_{t+1}^i,\ldots,\hat p_{t+L}^i)$ and $\tau^\ast=(p_{t+1}^{i,*},\ldots,p_{t+L}^{i,*})$ denote its generated and ground-truth future trajectories, with $\hat p_t^i=p_t^{i,*}=p_t^i$. The imitation loss combines stepwise position error and soft-DTW~\cite{cuturi2017soft}:
\begin{equation}
\begin{aligned}
\mathcal L_{\mathrm{imit}}(\tau,\tau^\ast)
&=\frac{\lambda_{\mathrm{pos}}}{LG}
   \sum_{j=1}^{L}\|\hat p_{t+j}^i-p_{t+j}^{i,*}\|_2\\
&\quad+\lambda_{\mathrm{dtw}}
   \operatorname{soft\text{-}DTW}_{\gamma}(\tau,\tau^\ast).
\end{aligned}
\label{eq:app_imitation}
\end{equation}
Soft-DTW computes a soft minimum over temporal alignment costs:
\begin{equation}
\begin{aligned}
\operatorname{soft\text{-}DTW}_{\gamma}(\tau,\tau^\ast)
&=-\gamma\log\sum_{\pi\in\Pi_{L,L}}
\exp\left(-\frac{C_\pi}{\gamma}\right),\\
C_\pi
&=\sum_{(a,b)\in\pi}
\delta\bigl(\hat p_{t+a}^i,p_{t+b}^{i,*}\bigr),
\end{aligned}
\label{eq:app_soft_dtw}
\end{equation}
where $\Pi_{L,L}$ contains alignment paths from $(1,1)$ to $(L,L)$ with steps $(1,0)$, $(0,1)$, or $(1,1)$. We use the Euclidean matching cost $\delta(x,y)=\|x-y\|_2/G$, equivalent to computing the L2 distance between coordinates normalized by $G$. The parameter $\gamma>0$ controls the smoothing. Stepwise error compares positions at corresponding times. Soft-DTW allows differences in timing while preserving temporal order. We use $\lambda_{\mathrm{pos}}=1$, $\lambda_{\mathrm{dtw}}=0.5$, and $\gamma=1$.

\paragraph{Kinematic supervision.}
We match normalized step lengths and their successive changes between generated and ground-truth trajectories. For $j=1,\ldots,L-1$, let
\begin{equation}
\begin{aligned}
\nu_{t,j}^{i,*}
&=\frac{\|p_{t+j+1}^{i,*}-p_{t+j}^{i,*}\|_2}{G},\\
\hat\nu_{t,j}^{i}
&=\frac{\|\hat p_{t+j+1}^{i}-\hat p_{t+j}^{i}\|_2}{G}.
\end{aligned}
\label{eq:app_step_lengths}
\end{equation}
At a fixed temporal sampling interval, these step lengths are proportional to pedestrian speeds. The kinematic loss is
\begin{equation}
\begin{aligned}
\mathcal L_{\mathrm{kin}}
&=\frac{1}{L-1}\sum_{j=1}^{L-1}
|\hat\nu_{t,j}^{i}-\nu_{t,j}^{i,*}|\\
&\quad+\frac{1}{L-2}\sum_{j=1}^{L-2}
\bigl|(\hat\nu_{t,j+1}^{i}-\hat\nu_{t,j}^{i})
-(\nu_{t,j+1}^{i,*}-\nu_{t,j}^{i,*})\bigr|.
\end{aligned}
\label{eq:app_kin}
\end{equation}
The loss uses step lengths between future positions and excludes the initial step from the observed position in both trajectories.

\paragraph{Adversarial supervision.}
The discriminator receives a sequence of $L$ normalized displacement vectors, including the first displacement from the observed position. With $\hat p_t^i=p_t^{i,*}=p_t^i$, define
\begin{equation}
\begin{aligned}
\Delta\tau_t^i
&=\left(\frac{\hat p_{t+j}^i-\hat p_{t+j-1}^i}{G}\right)_{j=1}^{L},\\
\Delta\tau_t^{i,*}
&=\left(\frac{p_{t+j}^{i,*}-p_{t+j-1}^{i,*}}{G}\right)_{j=1}^{L}.
\end{aligned}
\label{eq:app_displacement_sequences}
\end{equation}
Let $D$ denote the discriminator.
Using least-squares adversarial supervision~\cite{mao2017least}, we define
\begin{equation}
\begin{aligned}
\mathcal L_{\mathrm{adv}}
&=\bigl(D(\Delta\tau_t^i)-1\bigr)^2,\\
\mathcal L_D
&=\bigl(D(\Delta\tau_t^{i,*})-1\bigr)^2
+D(\Delta\tau_t^i)^2,\\
\mathcal L_{\mathrm{act}}
&=\mathcal L_{\mathrm{imit}}
+\lambda_{\mathrm{kin}}^{\mathrm a}\mathcal L_{\mathrm{kin}}
+\lambda_{\mathrm{adv}}\mathcal L_{\mathrm{adv}}.
\end{aligned}
\label{eq:app_adversarial_full}
\end{equation}
The discriminator minimizes $\mathcal L_D$ with targets of one for ground-truth motion and zero for generated motion. The actor minimizes $\mathcal L_{\mathrm{act}}$, whose adversarial term encourages generated motion to receive a discriminator score of one. We use $\lambda_{\mathrm{adv}}=0.3$.

\paragraph{Behavior optimization and gradient flow.} We train the actor, critic, and discriminator for 40 epochs with four trajectory windows per batch and $L=12$ rollout steps, averaging losses over valid pedestrians. The actor and critic use AdamW with learning rate $3\times10^{-4}$ and weight decay $10^{-5}$. The discriminator uses Adam with learning rate $2\times10^{-4}$ and $(\beta_1,\beta_2)=(0.5,0.9)$. Gradient norms are clipped at 1.0, and we perform two critic updates per actor update. The actor minimizes $\mathcal L_{\mathrm{act}}$, the discriminator minimizes $\mathcal L_D$, and the critic minimizes the listwise loss defined below. During actor training, $\phi_{\mathrm{rep}}$ remains frozen. Gradients pass through explicit position updates, while rasterized observations and extracted features are detached from the actor's computation graph. The collision term is used in critic supervision, and user-defined spatial costs are introduced only during planning.

\paragraph{Critic candidates and supervision.}
For each observed crowd state $s_t$ and pedestrian $i$, we sample $K$ displacements around the actor proposal $\bar a_t^i$. Candidate $k$ replaces only pedestrian $i$'s first displacement. Other pedestrians follow the actor at that step, and all pedestrians follow the actor from the updated states for the remaining $L-1$ steps. Each candidate therefore produces an $L$-step crowd rollout that includes surrounding pedestrians' predicted responses to the initial action. For candidate $k$, we compute the trajectory cost for pedestrian $i$ as
\begin{equation}
y_k=\mathcal L_{\mathrm{imit}}
+\lambda_{\mathrm{kin}}^{\mathrm c}\mathcal L_{\mathrm{kin}}
+\lambda_{\mathrm{col}}\mathcal L_{\mathrm{col}}.
\label{eq:app_critic_cost}
\end{equation}
The imitation and kinematic terms use the same definitions as in actor training, comparing pedestrian $i$'s simulated trajectory with the same ground-truth trajectory across candidates. Let $\hat p_{t+j}^{b,(k)}$ denote the position of pedestrian $b\in\mathcal A_t$ at step $j$ of candidate $k$'s rollout. The collision term counts neighbors within $r=4$ grid pixels of pedestrian $i$:
\begin{equation}
\mathcal L_{\mathrm{col}}
=\frac{1}{L}\sum_{j=1}^{L}
\sum_{b\in\mathcal A_t\setminus\{i\}}
\mathbf 1\bigl[
\|\hat p_{t+j}^{i,(k)}-\hat p_{t+j}^{b,(k)}\|_2<r
\bigr].
\label{eq:app_collision_cost}
\end{equation}
The collision term averages neighbor counts over time, so proximity
to multiple neighbors and repeated proximity both contribute.

\paragraph{Critic ranking loss.}
The critic observes only the first $H$ steps of each candidate rollout,
with $1\le H<L$, and predicts
$q_k=\phi_{\mathrm{critic}}(\hat s_{t+1:t+H}^{(k)},i)$.
Given the $L$-step costs $\mathbf y=(y_1,\ldots,y_K)$
and predicted scores $\mathbf q=(q_1,\ldots,q_K)$,
we use the listwise cross-entropy loss~\cite{cao2007learning}:
\begin{equation}
\begin{gathered}
\mathcal L_{\mathrm{critic}}=-\sum_{k=1}^{K}P_k\log Q_k,\\
P_k=\operatorname{softmax}(-\mathbf y/\tau_c)_k,\qquad
Q_k=\operatorname{softmax}(\mathbf q/\tau_c)_k.
\end{gathered}
\label{eq:app_listnet_full}
\end{equation}
We set the softmax temperature to $\tau_c=1$ and normalize over candidates
for the same state and pedestrian.
Lower-cost candidates receive higher target probabilities,
encouraging the critic to assign them higher scores.
By default, we use $L=12$ and $H=4$, so the critic learns to rank
candidates from four simulated steps using costs computed over
twelve steps. For the component ablations in Table~\ref{tab:ablation},
the critic is trained with $H=2$.
Ground-truth trajectories provide supervision through these training costs.
During planning, the critic receives only features from simulated crowd trajectories.

\paragraph{Loss-weight calibration.}
We calibrate the actor weight $\lambda_{\mathrm{kin}}^{\mathrm a}$ separately from the critic-target weights $\lambda_{\mathrm{kin}}^{\mathrm c}$ and $\lambda_{\mathrm{col}}$. The actor kinematic term is inactive for the first eight epochs. Its subsequent weight is 0.3 times the ratio of the mean actor-gradient norm of $\mathcal L_{\mathrm{imit}}+\lambda_{\mathrm{adv}}\mathcal L_{\mathrm{adv}}$ to that of $\mathcal L_{\mathrm{kin}}$. Each critic-target weight is 0.3 times the ratio of the mean standard deviation of imitation costs across candidates to that of the corresponding term. Here, 0.3 is a calibration factor rather than a fixed loss weight. Table~\ref{tab:hparams} summarizes the training and planning settings.

\begin{table}[!htbp]
\caption{\textbf{Training and planning settings.}
User-cost weights are reported with each control protocol.}
\label{tab:hparams}
\centering\small
\setlength{\tabcolsep}{4pt}
\renewcommand{\arraystretch}{1.08}
\begin{tabular}{@{}p{.57\linewidth}p{.34\linewidth}@{}}
\toprule
Setting & Value \\
\midrule
\multicolumn{2}{@{}l}{\textit{Prediction pretraining}}\\
Optimizer & AdamW \\
Learning rate / weight decay & $3\times10^{-4}$ / $10^{-5}$ \\
Learning-rate schedule / gradient clipping & Cosine / 1.0 \\
Epochs & 80 \\
Batch size / accumulation steps & 1 / 3 \\
Prediction horizon $L$ & 12 \\
Waypoint / auxiliary weights & $\lambda_{\mathrm{way}}=1$, $\lambda_{\mathrm{aux}}=0.5$ \\
Heatmap positive-class weight & 50 \\
\midrule
\multicolumn{2}{@{}l}{\textit{Behavior learning}}\\
Epochs / trajectory windows per batch & 40 / 4 \\
Rollout length & $L=12$ \\
Actor and critic optimizer & AdamW \\
Actor/critic learning rate / weight decay & $3\times10^{-4}$ / $10^{-5}$ \\
Discriminator optimizer & Adam \\
Discriminator learning rate / $(\beta_1,\beta_2)$ & $2\times10^{-4}$ / $(0.5,0.9)$ \\
Gradient clipping / critic updates per actor update & 1.0 / 2 \\
Position / soft-DTW weights & $\lambda_{\mathrm{pos}}=1$, $\lambda_{\mathrm{dtw}}=0.5$ \\
Soft-DTW smoothing / adversarial weight & $\gamma=1$, $\lambda_{\mathrm{adv}}=0.3$ \\
Actor kinematic weight $\lambda_{\mathrm{kin}}^{\mathrm a}$ & Gradient-norm calibrated \\
Actor kinematic warm-up / calibration factor & 8 epochs / 0.3 \\
Critic-target weights $\lambda_{\mathrm{kin}}^{\mathrm c}$, $\lambda_{\mathrm{col}}$ & Standard-deviation calibrated \\
Critic-target calibration factor & 0.3 for each term \\
Collision threshold / listwise temperature & $r=4$ grid pixels / $\tau_c=1$ \\
\midrule
\multicolumn{2}{@{}l}{\textit{CEM planning}}\\
Default candidates / iterations / horizon & $K=16$, $I=4$, $H=4$ \\
Elite fraction & Highest-scoring quarter \\
Initial standard deviation & $\sigma_{\mathrm{plan}}=1.5$ grid pixels \\
\bottomrule
\end{tabular}
\end{table}

\subsection{Planning, Goal Updates, and Execution}
\label{app:planning}
\paragraph{Local goals.}
In trajectory-generation benchmarks, the pretrained goal head supplies $g_t^i$ at the start of each $L$-step window. The goal coordinate remains fixed within that window, while its direction $u_t^i$ is updated from the pedestrian's current position. The goal-refresh interval $L$ is distinct from the planning horizon $H$. Actor training uses local goals sampled from ground-truth trajectories. During trajectory forecasting, the waypoint head is conditioned on a predicted endpoint. The Habitat integration uses episode waypoint routes, as described in Appendix~\ref{app:habitat}. These local goals are separate from the user-defined regions or targets in the spatial cost.

\paragraph{Candidate rollouts.}
For each candidate, pedestrian $i$ uses the sampled displacement at the first step, while the others use their actor proposals. All pedestrians then follow the actor from the updated simulated crowd states for the remaining $H-1$ steps. Positions, velocity histories, and interaction features are updated at each step, allowing nearby pedestrians to respond to the candidate action. The active set $\mathcal A_t$ remains fixed within each rollout.

\paragraph{Feature caching.}
Within each planning rollout, we cache U-Net feature maps computed from the static scene map and initial pedestrian heatmaps. At each step, we update interaction features and sample local scene features from the cached maps at the updated pedestrian positions. Because the cached maps encode initial pedestrian positions as well as scene geometry, this approximates the full feature recomputation used during actor training.

\paragraph{CEM action selection.}
For each pedestrian $i$ selected for planning, we use the cross-entropy method (CEM)~\cite{de2005tutorial} to search for its next displacement. We initialize an isotropic Gaussian with mean $\bar a_t^i$ and standard deviation $\sigma_{\mathrm{plan}}=1.5$ grid pixels. At each of $I$ iterations, we sample $K$ candidate displacements and simulate an $H$-step crowd rollout for each candidate. We rank candidates using
\begin{equation}
J_k=q_k-\lambda_{\mathrm{user}}c_k.
\label{eq:app_planning_score}
\end{equation}
Here, $\lambda_{\mathrm{user}}\ge0$ weights the spatial cost. The critic score $q_k$ evaluates pedestrian $i$'s motion, whereas $c_k$ evaluates all active pedestrians over all $H$ simulated steps, using the definitions in Appendix~\ref{app:control_metrics}. At each iteration, CEM updates the Gaussian mean and per-coordinate variances using the highest-scoring quarter of candidates ranked by $J_k$. After $I$ iterations, we use the final Gaussian mean as the planned displacement $a_t^i$. The default planning settings are $K=16$, $I=4$, and $H=4$,
retaining four elite candidates per iteration.
The avoidance evaluation in Table~\ref{tab:control} uses $I=8$.

\paragraph{Simultaneous execution.}
Every selected pedestrian plans independently from the same current state $s_t$, assuming actor-generated actions for the others in its candidate rollouts. After all decisions are made, we update positions simultaneously:
\begin{equation}
p_{t+1}^i =
\begin{cases}
p_t^i + a_t^i,      & \text{if selected for planning}, \\
p_t^i + \bar a_t^i, & \text{otherwise}.
\end{cases}
\label{eq:app_transition}
\end{equation}
Velocity histories are updated from the executed motion before replanning. This procedure accounts for predicted neighbor responses while planning each pedestrian's action independently. Responses imagined under actor-generated actions may therefore differ from those following the simultaneously executed planned actions. Algorithm~\ref{alg:ctrl_cwm} summarizes one simulation step.

\begin{algorithm}[t]
\caption{Ctrl-CWM planning and simultaneous execution}
\label{alg:ctrl_cwm}
\label{alg:step}
\footnotesize
\begin{algorithmic}[1]
\Require Current state $s_t$, pedestrians selected for planning, user cost,
weight $\lambda_{\mathrm{user}}$, and $K,I,H,\sigma_{\mathrm{plan}}$
\Ensure Updated crowd state $s_{t+1}$
\State Refresh local goals when due under the execution protocol
\For{each active pedestrian $i$}
  \State $z_t^i\gets\phi_{\mathrm{rep}}(s_t,i)$
  \State $\bar a_t^i\gets\phi_{\mathrm{actor}}(z_t^i,u_t^i,v_t^i)$
\EndFor
\For{each pedestrian $i$ selected for planning}
  \State Initialize Gaussian mean $\bar a_t^i$ and isotropic standard deviation $\sigma_{\mathrm{plan}}$
  \For{$m=1,\ldots,I$}
    \State Sample $K$ candidate next-step displacements
    \For{each candidate $k$}
      \State Copy $s_t$ with active set $\mathcal A_t$ fixed
        \State Cache initial U-Net features
      \State Replace only $i$'s first displacement with candidate $k$
      \State Apply actor proposals for all other pedestrians
      \State Update all imagined positions and velocity histories
      \For{$h=1,\ldots,H-1$}
        \State Update interaction features and local feature-sampling positions
        \State Generate all displacements with $\phi_{\mathrm{actor}}$
        \State Update all imagined positions and velocity histories
      \EndFor
      \State $q_k\gets\phi_{\mathrm{critic}}(\hat s_{t+1:t+H}^{(k)},i)$
      \State Compute $c_k$ over all active pedestrians and $H$ steps
      \State $J_k\gets q_k-\lambda_{\mathrm{user}}c_k$
    \EndFor
    \State Update Gaussian mean and per-coordinate variances using the highest-scoring quarter by $J_k$
  \EndFor
  \State Set $a_t^i$ to the final Gaussian mean
\EndFor
\State Apply all executed displacements using Eq.~\ref{eq:app_transition}
\State Update velocity histories to obtain $s_{t+1}$
\end{algorithmic}
\end{algorithm}

\paragraph{Generation and control.}
Actor-only generation directly executes $\bar a_t^i$ for every pedestrian. Full-model generation plans for all active pedestrians with $\lambda_{\mathrm{user}}=0$, retaining the learned critic in action selection. Run-time control introduces or updates the spatial cost without retraining the actor or critic. Habitat restricts planning to pedestrians near the robot. The emitter handles entries outside imagined rollouts, while Ctrl-CWM generates motion for the active crowd.

\subsection{Checkpoint Selection}
\label{app:ckpt}
We select the actor and critic as a pair using held-out checkpoint-selection data. Actor quality is measured by full-rollout position error. For critic quality, we compute Spearman correlation between predicted scores and negative candidate costs within each state--pedestrian candidate set and average these correlations over the selection data. Among checkpoints within $5\%$ of the lowest position error, we select the pair with the highest ranking correlation. The selected pairs achieve correlations of 0.60--0.76 across folds.

\section{Evaluation Protocols and Metrics}
\label{app:protocols}
\subsection{Evaluation Scope}
We evaluate trajectory generation and spatial control on ETH--UCY, SDD, and GCS. The main averages give equal weight to seven benchmark entries: five ETH--UCY folds, SDD, and GCS. Component and sensitivity analyses average the five ETH--UCY folds. Social-HM3D evaluates crowd responses to a moving robot with fixed navigation methods. Prediction-head evaluation separately measures forecasting accuracy. The user study evaluates perceived objective following and recognition of robot-directed behavior. Table~\ref{tab:protocol_summary} summarizes these settings.

Generation metrics and control compliance use their respective evaluation windows. Generation metrics in the main component ablations (Table~\ref{tab:ablation}) use full episodes, while avoidance compliance and planning-parameter analyses use post-activation windows. The user-cost-only variant removes the critic score from candidate ranking under avoidance control. The component ablations in Table~\ref{tab:ablation} use an avoidance margin of $1.5\times$ the zone radius, $I=4$ CEM iterations, and a critic trained with $H=2$, whereas Table~\ref{tab:control} uses a $3\times$ margin, $I=8$, and a critic trained with $H=4$.

\begin{table*}[t]
\caption{\textbf{Evaluation protocols.} Generation and control evaluations use the windows indicated below.}
\label{tab:protocol_summary}
\centering\small
\setlength{\tabcolsep}{4pt}
\renewcommand{\arraystretch}{1.15}
\begin{tabular}{@{}p{.23\linewidth}p{.39\linewidth}p{.26\linewidth}@{}}
\toprule
Evaluation & Setting and aggregation & Repeats and window \\
\midrule
Generation, Table~\ref{tab:gen_shared} & Two replayed arrival protocols. Mean across seven entries: five ETH--UCY folds, SDD, and GCS. & Full episodes. \\
Avoidance, Table~\ref{tab:control}
& Disc, rectangle, multiple zones. ORCA obstacles, TRACE guidance,
CrowdES constraints. Ctrl-CWM: $\lambda_{\mathrm{user}}=10$,
$3\times$ margin, $I=8$, critic trained with $H=4$.
& 10 trials per scene. Final two thirds. \\
Components, Table~\ref{tab:ablation}
& Diffusion arrivals over five folds. $1.5\times$ margin, $I=4$,
critic trained with $H=2$. User-cost-only ranks candidates by spatial
cost during avoidance control.
& Full episodes for generation. Post-activation window for control. \\
Planning, Fig.~\ref{fig:planning_ablation} & Five folds. $K$, $I$, and $H$ varied individually. & Post-activation window. \\
Additional components, Appendix~\ref{app:ablation} & Five folds. Disc avoidance, $\lambda_{\mathrm{user}}=10$. & 20 trials per scene. Final two thirds. \\
Weight/\allowbreak population sweeps, Appendix~\ref{app:control} & Five folds. Weights and emitter multipliers varied as reported. & One trial per configuration. Post-activation window. \\
Habitat, Appendix~\ref{app:habitat} & 100 scenes, 1{,}087 episodes. Fixed A$^{*}$ and ORCA planners and Falcon policy. & Complete robot-navigation episodes. \\
User study, Appendix~\ref{app:user_study} & 100 participants. Four paired comparisons and eight three-clip trials. & Substantive responses. Attention checks are separate. \\
Prediction heads, Appendix~\ref{app:prediction} & Eight observed, twelve predicted steps. Best-of-20 ADE/FDE. & Prediction windows, not controlled crowd episodes. \\
\bottomrule
\end{tabular}
\end{table*}

\paragraph{Shared arrivals.}
Within each generation protocol, ORCA, TRACE, CrowdES, and Ctrl-CWM receive identical replayed entries from the random-surface or diffusion emitter. The emitter specifies when and where pedestrians enter, allowing the comparison to focus on the motion produced after arrival. The evaluation covers full episodes, and the two arrival protocols are reported separately.

\paragraph{Run-time avoidance.}
Controlled and uncontrolled Ctrl-CWM runs share the first third of each episode. The controlled run then activates $\lambda_{\mathrm{user}}=10$, while its counterpart continues critic-guided generation with $\lambda_{\mathrm{user}}=0$. ORCA receives avoidance zones as obstacles, TRACE receives them as test-time guidance terms, and CrowdES receives them as map constraints. For each baseline, the controlled run activates its control mechanism from initialization and is compared with a corresponding uncontrolled run. All methods use identical arrivals, with 10 trials per scene evaluated over the final two thirds. The intervention timing therefore differs between Ctrl-CWM and the baselines. Attraction and sensitivity experiments use the weights and repeat counts reported in Appendix~\ref{app:control}.

\subsection{Generation Metrics}
\label{app:metrics}
We follow the released CrowdES evaluation implementation~\cite{bae2025crowdes} to compare generated trajectories with recorded crowd motion. Density, frequency, coverage, and population describe scene-level distributions. Kinematics, trajectory matching, diversity, and collisions describe agent-level outcomes. Lower values are preferred except for diversity.

\paragraph{Scene-level distributions.}
For the benchmark's $Q\times Q$ evaluation grid, with $Q=10$, let $n_{q,t}$ count agents in cell $q$, $u_{q,t}$ count distinct agent categories in that cell, and $N_t$ denote the scene population at time $t$. The statistics are
\begin{equation}
\begin{aligned}
 d_t &= \frac{1}{Q^2}\sum_{q=1}^{Q^2} n_{q,t},
 & f_t &= \frac{1}{Q^2}\sum_{q=1}^{Q^2} u_{q,t}, \\
 c_t &= \frac{1}{Q^2}\sum_{q=1}^{Q^2}\mathbf{1}[n_{q,t}>0],
 & n_t^{\mathrm{rel}} &= N_t/\overline N_{\mathrm{ref}}.
\end{aligned}
\label{eq:app_scene_statistics}
\end{equation}

These correspond to Dens., Freq., Cov., and Pop., respectively. The scalar $n_t^{\mathrm{rel}}$ denotes relative population, distinct from pedestrian position $p_t^i$. Here, $\overline N_{\mathrm{ref}}$ is the mean reference population over the frames used by the population metric, and both generated and reference counts use this same normalization. The released quadrat implementation fills empty sampled frames with zeros, while its population implementation uses frames containing agent records. Sampling at one-second intervals gives empirical distributions $P_m^{\mathrm{gen}}$ and $P_m^{\mathrm{ref}}$, compared using Earth Mover's Distance:
\begin{equation}
\begin{aligned}
E_m &= \operatorname{EMD}
\bigl(P_m^{\mathrm{gen}},P_m^{\mathrm{ref}}\bigr),\\
m &\in
\{\mathrm{Dens.},\mathrm{Freq.},\mathrm{Cov.},\mathrm{Pop.}\}.
\end{aligned}
\label{eq:app_scene_emd}
\end{equation}

This evaluation grid is separate from the model's $192\times192$ input grid. Freq. and Cov. coincide for a single agent category, as in ETH--UCY and GCS. SDD contains multiple categories, so these measures can differ.

\paragraph{Kinematics (Kinem.).}
The released evaluator computes each agent's mean speed, mean acceleration magnitude, total travel distance, and travel duration. It compares the resulting distributions across agents. For each component, both generated and reference values are divided by the reference mean before computing EMD. For the normalized distributions $\widetilde P_k$, the aggregate is
\begin{equation}
 \mathrm{Kinem.}=\frac{1}{|\mathcal K|}\sum_{k\in\mathcal K}
 \operatorname{EMD}\bigl(\widetilde P_k^{\mathrm{gen}},\widetilde P_k^{\mathrm{ref}}\bigr),
\label{eq:app_kinematics_metric}
\end{equation}
where $\mathcal K=\{\mathrm{speed},\mathrm{accel.},\mathrm{dist.},\mathrm{time}\}$.

\paragraph{Trajectory matching (DTW).}
Let $\mathcal T_{\mathrm{gen}}$ and $\mathcal T_{\mathrm{ref}}$ denote generated and reference trajectory sets.
The directed nearest-match distance is
\begin{equation}
d(\mathcal X,\mathcal Y)=\frac{1}{|\mathcal X|}
\sum_{\tau\in\mathcal X}\min_{\eta\in\mathcal Y}\operatorname{DTW}(\tau,\eta).
\label{eq:app_directed_dtw}
\end{equation}
The reported score averages both directions and normalizes by the evaluation frame rate:

\begin{equation}
 \mathrm{DTW}=\frac{
 d(\mathcal T_{\mathrm{gen}},\mathcal T_{\mathrm{ref}})
 +d(\mathcal T_{\mathrm{ref}},\mathcal T_{\mathrm{gen}})
 }{2f_{\mathrm{eval}}}.
\label{eq:app_dtw_metric}
\end{equation}

Here, $f_{\mathrm{eval}}$ is the frame rate of the trajectories used for DTW evaluation. Positions are expressed in meters. Following the released evaluator, DTW uses squared Euclidean matching costs and takes the square root of the minimum accumulated cost.

\paragraph{Diversity (Div.).}
For source set $\mathcal X$ and target trajectories
$\mathcal Y=\{\eta_1,\ldots,\eta_{|\mathcal Y|}\}$, let
$j(\tau)=\arg\min_{1\le b\le|\mathcal Y|}
\operatorname{DTW}(\tau,\eta_b)$.
The directed coverage is the fraction of target trajectories selected by at least one source trajectory,
\begin{equation}
 J(\mathcal X,\mathcal Y)=
 \frac{|\{j(\tau):\tau\in\mathcal X\}|}{|\mathcal Y|}.
 \label{eq:app_diversity_coverage}
\end{equation}
Div. averages $J(\mathcal T_{\mathrm{gen}},\mathcal T_{\mathrm{ref}})$ and
$J(\mathcal T_{\mathrm{ref}},\mathcal T_{\mathrm{gen}})$.
Higher values indicate broader coverage under nearest-trajectory matching.

\paragraph{Collision rate (Col.).}
For the active agent set $\mathcal A_t$, the reported collision percentage is
\begin{equation}
\mathrm{Col.}=100\,
\frac{\displaystyle\sum_t\sum_{i\in\mathcal A_t}
\mathbf 1\!\left[\min_{j\in\mathcal A_t\setminus\{i\}}
\|p_t^i-p_t^j\|_2<0.2\,\mathrm m\right]}
{\displaystyle\sum_t|\mathcal A_t|}.
\label{eq:app_collision_metric}
\end{equation}
Positions are converted to metric coordinates. The indicator is zero when no other agent is present. An agent contributes at most once per step, even when multiple neighbors are within the threshold. This differs from Habitat H-coll., which reports the percentage of episodes containing at least one robot--human collision.

\paragraph{Training and evaluation quantities.}
The kinematic training loss compares scalar step-length profiles and successive differences, whereas Kinem. compares distributions of four motion quantities. The critic's collision cost counts neighboring pedestrians over imagined steps, whereas Col. is an evaluation percentage. Similarly, the soft-DTW training term and the nearest-trajectory DTW benchmark metric serve different purposes. Control compliance is defined in Appendix~\ref{app:control_metrics}.

\section{Run-time Control Costs and Sensitivity}
\label{app:control}
\label{app:control_analysis}
\subsection{Spatial Costs and Compliance}
\label{app:control_metrics}
Prior work has explored test-time pedestrian control through waypoint and obstacle-avoidance objectives~\cite{rempe2023trace}. Here, we define spatial costs and occupancy-based compliance measures for user-specified regions in continuous crowd simulation.

\begin{figure}[!htbp]
\centering
\includegraphics[width=0.9\columnwidth]{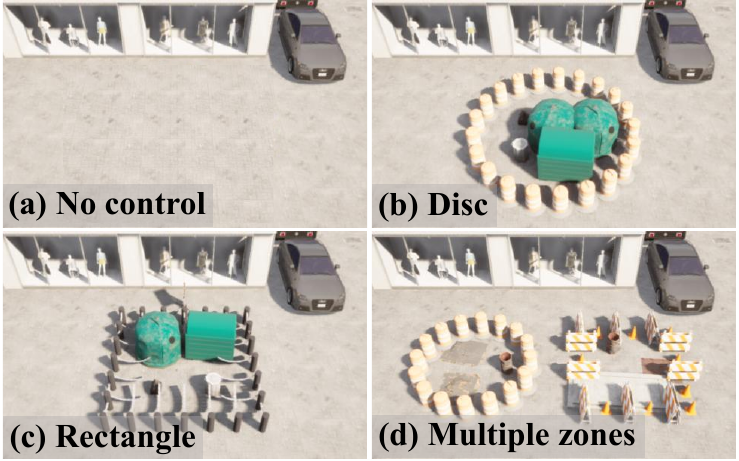}
\caption{\textbf{Control zones.}
(a) No control.
(b--d) Disc, rectangle, and multiple avoidance regions.
Zones are placed on the main pedestrian flow of each scene.}
\label{fig:control_zones}
\end{figure}

\paragraph{Control regions and planning costs.}
We place disc-shaped, rectangular, and multiple disjoint avoidance regions along the main pedestrian flow of each scene. These configurations test control across different region shapes and multiple separated locations. Figure~\ref{fig:control_zones} illustrates the configurations.

For a disc, $Z=\{p:\|p-z_c\|_2\le r_Z\}$, with center $z_c$ and radius $r_Z$.
For other configurations, $Z$ is the rectangle or union of disjoint regions.
The region radius is independent of the inter-agent collision threshold.
Write $\hat s^{(k)}=\hat s_{t+1:t+H}^{(k)}$ and let $\hat p_{t+h}^{b,(k)}$ denote
pedestrian $b$'s position in its $h$-th state. With $\mathcal A_t$ fixed within the rollout,
the spatial costs are

\begin{align}
c_{\mathrm{avoid}}(\hat s^{(k)})
&= \frac{1}{H|\mathcal A_t|} \sum_{h=1}^{H} \sum_{b\in\mathcal A_t}
\mathbf 1\bigl[ \hat p_{t+h}^{b,(k)} \in Z \bigr], \label{eq:app_avoid} \\
c_{\mathrm{attract}}(\hat s^{(k)})
&= \frac{1}{H|\mathcal A_t|} \sum_{h=1}^{H} \sum_{b\in\mathcal A_t}
\frac{\| \hat p_{t+h}^{b,(k)} - z_c \|_2}{G}. \label{eq:app_attract}
\end{align}

We set $c_k=c_{\mathrm{avoid}}(\hat s^{(k)})$ or $c_k=c_{\mathrm{attract}}(\hat s^{(k)})$
in Eq.~\ref{eq:app_planning_score}, according to the user objective. Both costs average
over all active pedestrians and all $H$ steps. Avoidance penalizes occupancy inside $Z$, while attraction penalizes distance to the target center. The distance cost distinguishes candidates even when all predicted positions remain outside the target region. The planner can also use other distance-based costs, for example to discourage pedestrians from coming too close to an object or encourage them to stay within a specified distance of a target.
Changing the zone or target updates the planning objective without retraining the actor or critic.
The static scene map remains unchanged for Ctrl-CWM control. In Habitat, the zone
center is updated from the robot position at each planning call
(Appendix~\ref{app:habitat}).

\paragraph{Region occupancy.}
For run $r\in\{\mathrm{cmd},\mathrm{free}\}$, let $\mathcal A_t^r$ be the active agents
and $p_t^{i,r}$ their positions.
Over evaluation window $\mathcal W$, occupancy is

\begin{equation}
 \mathrm{occ}_r=
 \frac{\displaystyle\sum_{t\in\mathcal W}\sum_{i\in\mathcal A_t^r}
       \mathbf{1}[p_t^{i,r}\in Z]}
      {\displaystyle\sum_{t\in\mathcal W}|\mathcal A_t^r|}.
\label{eq:app_region_occupancy}
\end{equation}

The subscripts denote runs with and without the control objective. We define compliance as the occupancy improvement relative to the uncontrolled run, normalized by the maximum possible improvement from that baseline:

\begin{equation}
\begin{aligned}
 C_{\mathrm{avoid}}
 &=1-\frac{\mathrm{occ}_{\mathrm{cmd}}}{\mathrm{occ}_{\mathrm{free}}}, \\
 C_{\mathrm{attract}}
 &=\frac{\mathrm{occ}_{\mathrm{cmd}}-\mathrm{occ}_{\mathrm{free}}}
         {1-\mathrm{occ}_{\mathrm{free}}}.
\end{aligned}
\label{eq:app_compliance}
\end{equation}

Avoidance requires $\mathrm{occ}_{\mathrm{free}}>0$, and attraction requires
$\mathrm{occ}_{\mathrm{free}}<1$.
Both scores are zero for unchanged occupancy.
Avoidance reaches one when controlled occupancy is zero, and attraction reaches one when
all evaluated agent steps lie in the target region.
Negative values indicate a change opposite to the objective.
These are region-occupancy measures, so we report motion statistics alongside compliance.

\paragraph{Paired sensitivity protocol.}
Weight and emitter-population sweeps use five ETH--UCY folds and one trial per configuration.
Each pair shares its first third, arrivals, and region geometry.
After activation, one run receives the user cost and the other continues with
$\lambda_{\mathrm{user}}=0$.
Changes in motion metrics are controlled minus uncontrolled values over the post-activation window.
Collision-rate changes are reported in percentage points (pp).
These one-trial sweeps describe sensitivity to the tested settings.

\subsection{Effect of the User-Cost Weight}
\label{app:cost_weight}
Avoidance and attraction use different cost functions and scales.
We therefore analyze their weights separately.

\paragraph{Avoidance.}
Table~\ref{tab:avoid_sweep} shows that average compliance increases from 0.607 at
$\lambda_{\mathrm{user}}=5$ to 0.863 at 10 and 0.985 at 25.
Further increases to 50 and 100 yield averages of 0.968 and 0.979.
The largest average increase in kinematic error across these settings is 0.099.
At weights of at least 25, some ETH, HOTEL, and UNIV rollouts slow or stop near the region.
Consequently, increased compliance can be accompanied by reduced progress as well as rerouting.

\begin{table}[!htbp]
\caption{\textbf{Avoidance weight sweep.} Five folds, one trial per configuration, post-activation evaluation. Kinematic changes are controlled minus uncontrolled. $\lambda$ is $\lambda_{\mathrm{user}}$.}
\label{tab:avoid_sweep}
\centering\small
\setlength{\tabcolsep}{4pt}
\renewcommand{\arraystretch}{1.14}
\begin{tabular*}{\linewidth}{@{\extracolsep{\fill}}lrrrrr@{}}
\toprule
\multicolumn{6}{@{}l}{\textit{Avoidance compliance $\uparrow$}} \\
Fold & $\lambda=5$ & $\lambda=10$ & $\lambda=25$ & $\lambda=50$ & $\lambda=100$ \\
\midrule
ETH & 0.657 & 0.949 & 0.987 & 0.955 & 0.997 \\
HOTEL & 0.661 & 0.823 & 0.990 & 0.964 & 0.992 \\
UNIV & 0.694 & 0.931 & 1.000 & 1.000 & 1.000 \\
ZARA1 & 0.422 & 0.759 & 0.963 & 0.973 & 0.938 \\
ZARA2 & 0.600 & 0.855 & 0.986 & 0.949 & 0.967 \\
\midrule
AVG & 0.607 & 0.863 & 0.985 & 0.968 & 0.979 \\
\midrule
\multicolumn{6}{@{}l}{\textit{$\Delta$Kinem. (controlled minus uncontrolled)}} \\
Fold & $\lambda=5$ & $\lambda=10$ & $\lambda=25$ & $\lambda=50$ & $\lambda=100$ \\
\midrule
ETH & $+0.006$ & $+0.058$ & $+0.042$ & $+0.040$ & $+0.032$ \\
HOTEL & $+0.143$ & $+0.165$ & $+0.157$ & $+0.157$ & $+0.158$ \\
UNIV & $-0.017$ & $-0.029$ & $-0.013$ & $+0.000$ & $-0.018$ \\
ZARA1 & $+0.037$ & $+0.032$ & $+0.173$ & $+0.186$ & $+0.127$ \\
ZARA2 & $+0.074$ & $+0.090$ & $+0.117$ & $+0.113$ & $+0.127$ \\
\midrule
AVG & $+0.049$ & $+0.063$ & $+0.095$ & $+0.099$ & $+0.085$ \\
\bottomrule
\end{tabular*}
\end{table}

\paragraph{Attraction.}
Average attraction compliance increases from 0.381 at $\lambda_{\mathrm{user}}=1.5$
to 0.645 at 3.0 and 0.951 at 5.0 (Table~\ref{tab:attract_sweep}).
The corresponding changes in motion statistics show the cost of stronger attraction
(Table~\ref{tab:attract_cost}).
At weight 3.0, mean kinematic error increases by 6.44 and collision rate by 2.8 pp.
At weight 5.0, these increases reach 7.76 and 26.5 pp.
Thus, higher target occupancy comes with substantial changes in motion statistics and collisions.

\begin{table}[!htbp]
\caption{\textbf{Attraction weight sweep.} Avoidance and attraction weights have different cost scales. One trial is used per configuration. AVG is calculated from the displayed fold values.}
\label{tab:attract_sweep}
\centering\small
\setlength{\tabcolsep}{4pt}
\renewcommand{\arraystretch}{1.14}
\begin{tabular*}{\linewidth}{@{\extracolsep{\fill}}lrrr@{}}
\toprule
Fold & $\lambda_{\mathrm{user}}=1.5$ & $\lambda_{\mathrm{user}}=3.0$ & $\lambda_{\mathrm{user}}=5.0$ \\
\midrule
ETH & 0.480 & 0.674 & 0.977 \\
HOTEL & 0.406 & 0.580 & 0.907 \\
UNIV & 0.244 & 0.670 & 0.960 \\
ZARA1 & 0.390 & 0.572 & 0.930 \\
ZARA2 & 0.386 & 0.727 & 0.983 \\
\midrule
AVG & 0.381 & 0.645 & 0.951 \\
\bottomrule
\end{tabular*}
\end{table}

\begin{table}[!htbp]
\caption{\textbf{Motion changes under attraction.} Post-activation controlled minus uncontrolled values under the attraction protocol. Collision changes are percentage points (pp), not relative percentages.}
\label{tab:attract_cost}
\centering\small
\setlength{\tabcolsep}{4pt}
\renewcommand{\arraystretch}{1.14}
\begin{tabular*}{\linewidth}{@{\extracolsep{\fill}}lrrrr@{}}
\toprule
 & \multicolumn{2}{c}{$\lambda_{\mathrm{user}}=3.0$} & \multicolumn{2}{c}{$\lambda_{\mathrm{user}}=5.0$} \\
\cmidrule(lr){2-3}\cmidrule(lr){4-5}
Fold & $\Delta$Kinem. & $\Delta$Col. (pp) & $\Delta$Kinem. & $\Delta$Col. (pp) \\
\midrule
ETH & $+13.92$ & $+11.2$ & $+13.78$ & $+37.9$ \\
HOTEL & $+5.41$ & $-0.4$ & $+10.66$ & $+22.2$ \\
UNIV & $+3.14$ & $+3.0$ & $+3.45$ & $+32.8$ \\
ZARA1 & $+4.09$ & $-0.5$ & $+4.98$ & $+15.6$ \\
ZARA2 & $+5.63$ & $+0.7$ & $+5.93$ & $+24.1$ \\
\midrule
AVG & $+6.44$ & $+2.8$ & $+7.76$ & $+26.5$ \\
\bottomrule
\end{tabular*}
\end{table}

\subsection{Sensitivity to the Emitter Population}
\label{app:density}
We vary the emitter population multiplier over $0.5$, $1$, and $2$ at avoidance weights 10 and 25.
These multipliers adjust the emission setting rather than directly specifying realized crowd density.
At weight 25, average compliance ranges from 0.928 to 0.985, and every fold achieves at least 0.847
(Table~\ref{tab:density}).
Collision-rate changes have magnitude at most 0.6 pp.
At weight 10, average compliance decreases from 0.863 at multiplier 1 to 0.701 at multiplier 2.
Collision-rate changes remain within 0.5 pp except for ETH at multiplier $0.5$, which increases by 9.5 pp.
The results indicate that emission conditions and control strength jointly influence the observed outcomes.

\begin{table}[!htbp]
\caption{\textbf{Emitter-population sensitivity.} Compliance and post-activation collision-rate changes (pp) at each emission multiplier. One trial is used per configuration. Values are rounded to the displayed precision.}
\label{tab:density}
\centering\small
\setlength{\tabcolsep}{3pt}
\renewcommand{\arraystretch}{1.14}
\begin{tabular*}{\linewidth}{@{\extracolsep{\fill}}lrrrrrr@{}}
\toprule
\multicolumn{7}{@{}l}{$\lambda_{\mathrm{user}}=10$} \\
 & \multicolumn{2}{c}{$\times 0.5$} & \multicolumn{2}{c}{$\times 1$} & \multicolumn{2}{c}{$\times 2$} \\
\cmidrule(lr){2-3}\cmidrule(lr){4-5}\cmidrule(lr){6-7}
Fold & Comp. $\uparrow$ & $\Delta$Col. & Comp. $\uparrow$ & $\Delta$Col. & Comp. $\uparrow$ & $\Delta$Col. \\
\midrule
ETH & 0.928 & $+9.5$ & 0.949 & $-0.0$ & 0.836 & $-0.1$ \\
HOTEL & 0.843 & $+0.0$ & 0.823 & $+0.0$ & 0.632 & $+0.0$ \\
UNIV & 1.000 & $-0.0$ & 0.931 & $-0.0$ & 0.732 & $+0.0$ \\
ZARA1 & 0.651 & $-0.2$ & 0.759 & $-0.2$ & 0.744 & $+0.5$ \\
ZARA2 & 0.683 & $-0.1$ & 0.855 & $-0.1$ & 0.563 & $-0.1$ \\
\midrule
AVG & 0.821 & $+1.8$ & 0.863 & $-0.1$ & 0.701 & $+0.1$ \\
\midrule
\multicolumn{7}{@{}l}{$\lambda_{\mathrm{user}}=25$} \\
 & \multicolumn{2}{c}{$\times 0.5$} & \multicolumn{2}{c}{$\times 1$} & \multicolumn{2}{c}{$\times 2$} \\
\cmidrule(lr){2-3}\cmidrule(lr){4-5}\cmidrule(lr){6-7}
Fold & Comp. $\uparrow$ & $\Delta$Col. & Comp. $\uparrow$ & $\Delta$Col. & Comp. $\uparrow$ & $\Delta$Col. \\
\midrule
ETH & 0.964 & $+0.1$ & 0.987 & $+0.1$ & 0.976 & $-0.1$ \\
HOTEL & 0.931 & $-0.1$ & 0.990 & $+0.0$ & 0.896 & $+0.1$ \\
UNIV & 1.000 & $+0.2$ & 1.000 & $-0.0$ & 1.000 & $+0.2$ \\
ZARA1 & 0.943 & $-0.2$ & 0.963 & $-0.1$ & 0.920 & $+0.6$ \\
ZARA2 & 0.913 & $-0.1$ & 0.986 & $-0.1$ & 0.847 & $-0.0$ \\
\midrule
AVG & 0.950 & $-0.0$ & 0.985 & $-0.0$ & 0.928 & $+0.2$ \\
\bottomrule
\end{tabular*}
\end{table}

\section{Robot-Reactive Crowd Control in Habitat 3.0}
\label{app:habitat}
\subsection{Benchmark and Crowd Integration}
We use Social-HM3D~\cite{gong2025cognition}, built on Habitat 3.0~\cite{puig2024habitat}, to evaluate control of pedestrians around a moving robot. The quantitative and qualitative results are reported in Sec.~\ref{sec:habitat}, Table~\ref{tab:nav_habitat}, and Fig.~\ref{fig:habitat_control}. Here, we detail the trajectory-to-humanoid interface and pedestrian-response measures. The evaluation uses all 1{,}087 validation episodes across 100 scenes. A$^{*}$ and ORCA planners and the released Falcon policy remain fixed across crowd conditions. Thus, the experiment tests transferred crowd controllability and its navigation effects, without training or modifying the robot navigation methods.

\paragraph{Transfer from pedestrian data.}
We transfer the model trained on GCS, as described in Sec.~\ref{sec:habitat}, without retraining or fine-tuning on Habitat data. HM3D floor plans are rasterized at the GCS map scale. Pedestrians retain their episode waypoints and nominal walking speeds of 0.8--1.2 times the robot's reference speed. Their nominal routes connect consecutive waypoints using navmesh shortest paths. The humanoid controller converts planned pedestrian positions into full-body movement. Thus, Ctrl-CWM controls trajectories, while Habitat provides their embodied execution.

\paragraph{Moving control objective.}
Ctrl-CWM replans pedestrian motion every 0.5\,s using the robot's current position. Only pedestrians within 6\,m of the robot are replanned. The control region is a disc of radius 1.5\,m centered on the robot and updated at each replanning call. We compare attraction with $\lambda_{\mathrm{user}}=10$, no user objective with $\lambda_{\mathrm{user}}=0$, and avoidance with $\lambda_{\mathrm{user}}=25$. All three conditions retain the actor, critic, and planning procedure. Replanned motion is blended into the nominal waypoint route, and the humanoid controller walks toward the resulting position in closed loop. Pedestrians beyond the replanning radius follow their nominal routes. The original benchmark humanoids provide a separate waypoint-following reference condition. The replanning interval is an execution setting and is distinct from the computation time per planning call.

\subsection{Navigation and Pedestrian-response Metrics}
\label{app:habitat_metrics}
We retain the benchmark's navigation measures and termination rules. Success rate (SR) is the fraction of successful episodes. Success weighted by path length (SPL) combines task completion with route efficiency. Personal-space compliance (PSC) follows the benchmark's criterion with a distance threshold of 1.0\,m. H-coll. is the fraction of episodes with at least one robot--human collision. Steps is the mean episode length, including successful and unsuccessful episodes. A shorter episode can therefore reflect either faster completion or earlier failure.

We additionally report control-zone occupancy (Zone), the fraction of
replanned pedestrian steps ending within the 1.5\,m robot-centered disc. Unlike $C_{\mathrm{avoid}}$ and $C_{\mathrm{attract}}$, Zone reports occupancy directly without normalization against a no-objective run. The zone moves with the robot, and its desired occupancy direction depends on the control objective.

\section{User Study}
\label{app:user_study}
\subsection{Purpose and Participants}
The study evaluates whether people recognize the effects of crowd control in two settings.
In Part 1, participants completed four paired-video comparison trials, selecting which crowd better followed a stated objective.
In Part 2, participants completed eight trials featuring pedestrian responses to a moving robot in Habitat.
Each trial contained three videos that participants labeled independently, yielding 24 labels per participant.
Both tasks assess perceived controllability, complementing the occupancy and navigation measurements in the main paper.

We recruited 100 participants through CloudResearch Connect~\cite{hartman2023introducing} for a desktop-only study. Participants were offered \$2.00 for an advertised duration of seven minutes. We did not seek ethics review because the task involved only viewing simulated videos and selecting responses, and we collected no names or contact information. Each part included one additional attention check, for two checks in total. The reported results use all 100 participants, regardless of attention-check performance. Attention-check responses are excluded from the task scores.

\begin{figure*}[t]
    \centering
    \includegraphics[width=\textwidth]{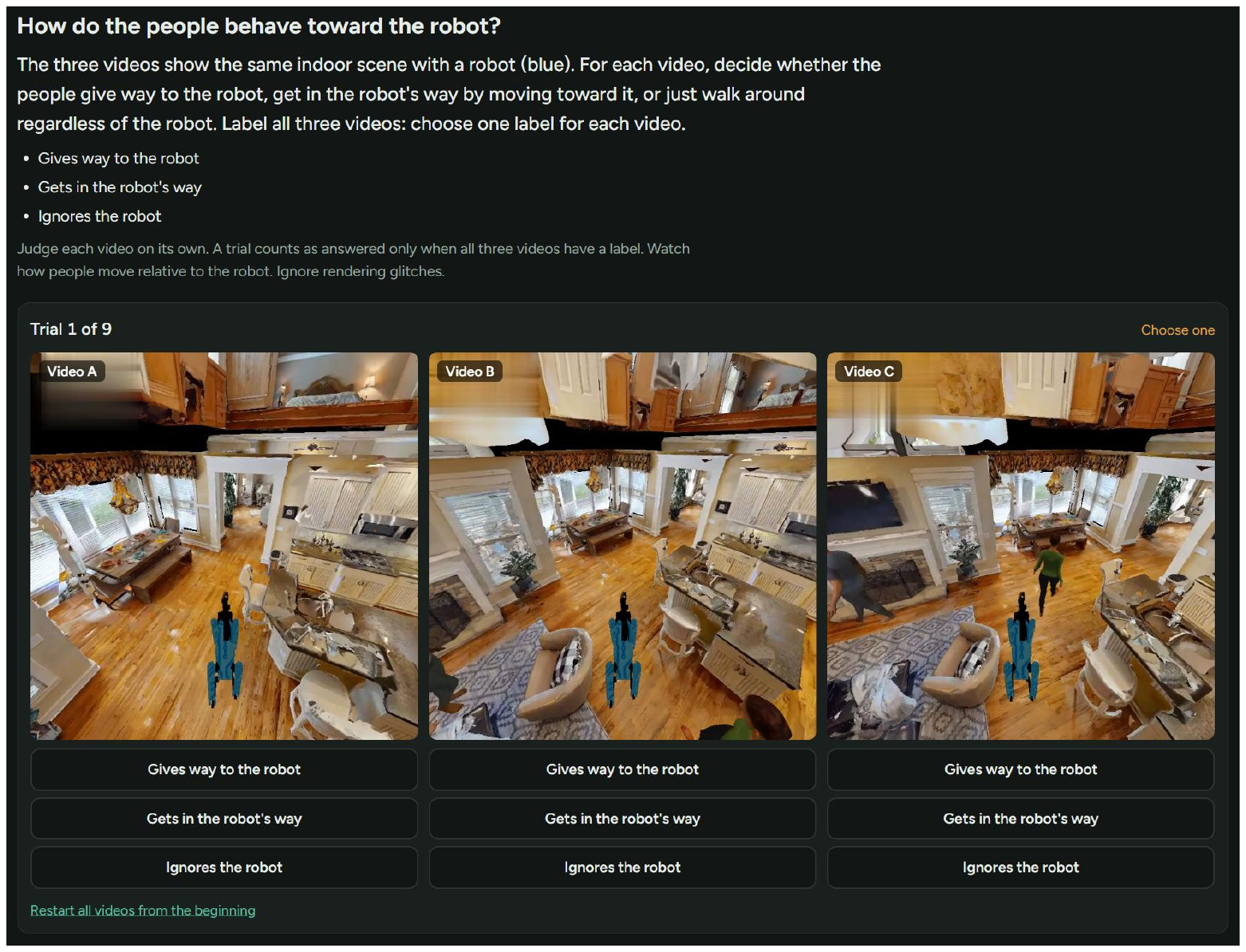}
    \caption{\textbf{User-study interface for robot-directed behavior
    recognition (Part 2).}
    Participants independently labeled pedestrian behavior in three
    videos as giving way to the robot, getting in the robot's way,
    or ignoring the robot.}
    \label{fig:user_study_interface}
\end{figure*}

\subsection{Stimuli and Tasks}
\paragraph{Scene-level objective following.}
Participants viewed paired videos of Ctrl-CWM and CrowdES crowds in four scenes. The attraction scenarios reflect observations of people spending time near fountains and gathering to watch performances in public plazas~\cite{carr1992public}. For the fountain and stage comparisons, the CrowdES videos were generated by modifying the obstacle map and rerunning the simulation. Each pair displayed a stated objective above the videos and asked, ``Which crowd is better controlled?'' Participants selected the video whose crowd more clearly followed the objective. Method names were hidden. Table~\ref{tab:user_study_objectives} gives the scene descriptions and displayed objectives. Videos were presented side by side at $960\times540$ resolution and ranged from 12 to 78 seconds. Each participant completed four substantive comparisons.

\begin{table*}[t]
\centering\small
\caption{\textbf{Scene-level user-study stimuli.}
Each trial compares Ctrl-CWM and CrowdES under the displayed objective.
Scene identifiers are retained for matching the stimuli to the study materials.}
\label{tab:user_study_objectives}
\setlength{\tabcolsep}{5pt}
\renewcommand{\arraystretch}{1.12}
\begin{tabular}{@{}lp{.24\linewidth}p{.64\linewidth}@{}}
\toprule
Scene & Event or target & Displayed objective \\
\midrule
A05 & Collision at a crossing & React to the collision (crash) that happens at the crossing. \\
A11 & Parked vehicle & Walk around and keep away from the parked vehicle. \\
A14 & Fountain & Be drawn toward the fountain. \\
A17 & Stage & Be drawn toward the stage. \\
\bottomrule
\end{tabular}
\end{table*}

\paragraph{Robot-directed behavior recognition.}
Participants viewed three clips for each of eight Habitat episodes across four indoor scenes.
The episodes were C05, C06, C08, C10, C11, C12, C16, and C17.
Each triplet contained Ctrl-CWM crowds generated under avoidance, attraction, and
no user objective (None). In this task, None is Ctrl-CWM without a user objective,
not the original benchmark humanoids used as a separate navigation reference.
Clips were presented side by side at $768\times768$ resolution and lasted 4--18 seconds.
The robot was shown in blue.
Figure~\ref{fig:user_study_interface} shows an example trial interface.

Participants labeled every clip independently using the three descriptions in
Table~\ref{tab:user_study_labels}. Labels could be reused within a triplet.
The task therefore did not require a one-to-one assignment between clips and labels.
Each participant provided 24 substantive labels, eight for each condition.

\begin{table}[t]
\centering\small
\caption{\textbf{Robot-directed behavior labels.}
All three conditions use Ctrl-CWM. The label associated with each condition is
used to calculate behavior-label agreement.}
\label{tab:user_study_labels}
\setlength{\tabcolsep}{4pt}
\renewcommand{\arraystretch}{1.12}
\begin{tabular}{@{}p{.16\linewidth}p{.35\linewidth}p{.39\linewidth}@{}}
\toprule
Mode & Spatial objective & Intended label \\
\midrule
Avoid & Avoid the robot-centered zone & Gives way to the robot \\
Attract & Approach the robot-centered zone & Gets in the robot's way \\
None & No user objective & Ignores the robot \\
\bottomrule
\end{tabular}
 
\end{table}

\begin{table}[!htbp]
\caption{\textbf{Interaction-encoder ablation.} Disc avoidance with $\lambda_{\mathrm{user}}=10$ and 20 trials per scene. AVG is calculated from the displayed fold values.}
\label{tab:interaction_ablation}
\centering\small
\setlength{\tabcolsep}{4pt}
\renewcommand{\arraystretch}{1.14}
\begin{tabular*}{\linewidth}{@{\extracolsep{\fill}}lrrrr@{}}
\toprule
 & \multicolumn{2}{c}{Full model} & \multicolumn{2}{c}{w/o interaction} \\
\cmidrule(lr){2-3}\cmidrule(lr){4-5}
Fold & Col. (\%) $\downarrow$ & Comp. $\uparrow$ & Col. (\%) $\downarrow$ & Comp. $\uparrow$ \\
\midrule
ETH & 0.458 & 0.932 & 20.32\phantom{0} & 0.950 \\
HOTEL & 1.522 & 0.992 & 1.292 & 0.497 \\
UNIV & 0.766 & 0.882 & 1.133 & 0.468 \\
ZARA1 & 3.508 & 0.839 & 0.868 & 0.466 \\
ZARA2 & 2.033 & 0.754 & 1.062 & 0.510 \\
\midrule
AVG & 1.657 & 0.880 & 4.935 & 0.578 \\
\bottomrule
\end{tabular*}
 
\end{table}

\begin{table}[!htbp]
\caption{\textbf{Auxiliary-head ablation.} Avoidance compliance under the same disc-control protocol. AVG is calculated from the displayed fold values.}
\label{tab:dynamics_ablation}
\centering\small
\setlength{\tabcolsep}{4pt}
\renewcommand{\arraystretch}{1.14}
\begin{tabular*}{\linewidth}{@{\extracolsep{\fill}}lrr@{}}
\toprule
Fold & Full model & w/o auxiliary head \\
\midrule
ETH & 0.932 & 0.880 \\
HOTEL & 0.992 & 0.215 \\
UNIV & 0.882 & 0.245 \\
ZARA1 & 0.839 & 0.880 \\
ZARA2 & 0.754 & 0.699 \\
\midrule
AVG & 0.880 & 0.584 \\
\bottomrule
\end{tabular*}
 
\end{table}

\begin{table}[!htbp]
\caption{\textbf{Critic-objective comparison.} Five-fold means under the 20-trial disc-control protocol.}
\label{tab:component_ablation}
\centering\small
\setlength{\tabcolsep}{3pt}
\renewcommand{\arraystretch}{1.14}
\begin{tabular*}{\linewidth}{@{\extracolsep{\fill}}lrrrr@{}}
\toprule
Variant & Kinem. $\downarrow$ & DTW $\downarrow$ & Col. (\%) $\downarrow$ & Comp. $\uparrow$ \\
\midrule
Pointwise critic & 0.433 & 0.994 & 2.454 & 0.870 \\
\textbf{Full model (listwise)} & \textbf{0.404} & \textbf{0.985} & \textbf{1.657} & \textbf{0.880} \\
\bottomrule
\end{tabular*}
 
\end{table}

\begin{table*}[t]
  \caption{\textbf{Trajectory prediction on ETH--UCY, SDD, and GCS.}
  Best-of-20 ADE/FDE is reported in meters for ETH--UCY and pixels for SDD and GCS.
  AVG is the mean over the five ETH--UCY folds.
  \textbf{Bold}/\underline{underline}: best/second-best distinct displayed values, with ties marked equally. Baseline results marked with $\dagger$ are obtained from our implementations. The remaining baseline results are taken from prior publications. N/A denotes unreported results.}
  \label{tab:pred_all}
  \centering
  \resizebox{0.85\linewidth}{!}{%
  \begin{tabular}{l|ccccc|c|c|c}
    \toprule
    Model & ETH & HOTEL & UNIV & ZARA1 & ZARA2 & AVG & SDD & GCS \\
    \midrule
    AgentFormer~\cite{yuan2021agentformer}  & 0.46/0.80 & 0.14/0.22 & 0.25/0.45 & 0.18/0.30 & 0.14/0.24 & 0.23/0.40 & 8.7/14.9 & 10.2/16.9 \\
    MID~\cite{gu2022stochastic}  & 0.57/0.93 & 0.21/0.33 & 0.29/0.55 & 0.28/0.50 & 0.20/0.37 & 0.31/0.54 & 7.6/14.3 & 10.7/18.2 \\
    EqMotion~\cite{xu2023eqmotion}  & 0.40/0.61 & \underline{0.12}/0.18 & 0.23/0.43 & 0.18/0.32 & \underline{0.13}/0.23 & \underline{0.21}/0.35 & 7.9/11.9 & 7.6/13.1 \\
    MART~\cite{lee2024mart}  & \underline{0.35}/0.47 & 0.14/0.22 & 0.25/0.45 & 0.17/\underline{0.29} & \underline{0.13}/\underline{0.22} & \underline{0.21}/0.33 & \textbf{7.4}/\underline{11.8} & 10.6/14.1 \\
    LMTraj~\cite{bae2024can}  & 0.41/0.51 & \underline{0.12}/\underline{0.16} & \underline{0.22}/\underline{0.34} & 0.20/0.32 & 0.17/0.27 & 0.22/0.32 & 7.8/\textbf{10.1} & \underline{7.1}/\textbf{9.6} \\
    MoFlow~\cite{fu2025moflow}  & 0.40/0.57 & \textbf{0.11}/0.17 & 0.23/0.39 & \textbf{0.15}/\textbf{0.26} & \textbf{0.12}/\underline{0.22} & \textbf{0.20}/0.32 & \underline{7.5}/12.0 & 9.1/11.6$^{\dagger}$ \\
    GUIDE-CoT~\cite{kim2025guide}  & 0.38/0.43 & 0.13/\textbf{0.15} & 0.34/0.48 & 0.19/\underline{0.29} & 0.16/\textbf{0.21} & 0.24/\underline{0.31} & \textcolor{gray}{N/A} & \textcolor{gray}{N/A} \\
    ViTE~\cite{li2026vite}  & \underline{0.35}/0.49 & \textbf{0.11}/0.17 & 0.23/0.42 & 0.18/0.30 & \underline{0.13}/\underline{0.22} & \textbf{0.20}/0.32 & \textbf{7.4}/11.9 & \textcolor{gray}{N/A} \\
    W2W~\cite{xu2026w2w}  & \underline{0.35}/\textbf{0.41} & \underline{0.12}/\textbf{0.15} & \textbf{0.20}/\textbf{0.32} & 0.19/\underline{0.29} & 0.17/0.26 & \underline{0.21}/\textbf{0.29} & \textbf{7.4}/\textbf{10.1} & \textcolor{gray}{N/A} \\
    \midrule
    \rowcolor{gray!20}
    \textbf{Ctrl-CWM}  & \textbf{0.34}/\underline{0.42} & 0.15/0.20 & 0.26/0.43 & \underline{0.16}/\textbf{0.26} & 0.14/0.24 & \underline{0.21}/\underline{0.31} & 7.8/11.9 & \textbf{5.9}/\underline{10.3} \\
    \bottomrule
    
  \end{tabular}}
  \vspace{-3mm}
\end{table*}

\subsection{Presentation and Response Controls}
\paragraph{Blinding and randomization.}
Method identities were not disclosed. Video filenames used random tokens to avoid
revealing the generating condition. For each participant, trial order within each
part and video positions within each trial were randomized. These assignments were
logged to recover the condition associated with each response.

\paragraph{Viewing requirements.}
Response buttons were enabled only after every video in a trial had played for at
least five seconds, or to completion for a shorter clip. Only the visible trial
played automatically, and clicking a video enabled full-screen viewing.
This requirement imposed a minimum viewing time without requiring complete playback
of every longer video.

\paragraph{Attention checks.}
Two additional trials assessed instruction following.
The Part 1 check showed two identical videos and explicitly instructed participants
to select Video B. The Part 2 check showed three identical videos and instructed
participants to assign ``Gets in the robot's way'' to A, ``Ignores the robot'' to B,
and ``Gives way to the robot'' to C.
These are instruction checks, separate from objective-following and behavior-recognition judgments.

\paragraph{Response collection.}
The study ran on a static website hosted on Netlify.
Responses were recorded through Google Apps Script in Google Sheets, and scoring
was performed server-side with the answer key omitted from the public page.
Connect identifiers were checked for duplicate submissions.
Browser-local storage blocked repeat submissions from the same browser.
This browser-level restriction did not prevent re-entry from another browser or device.

\subsection{Scoring and Interpretation}
Part 1 reports the percentage of substantive comparisons in which Ctrl-CWM was selected.
Part 2 reports the percentage of labels agreeing with the intended description,
calculated separately for Avoid, Attract, and None. A response is counted as agreement
only when its label matches the corresponding condition in Table~\ref{tab:user_study_labels}.
Attention-check trials do not contribute to either measure.

The aggregate results are reported in Table~\ref{tab:user_study}.
The paired comparisons assess whether the specified objective is visually apparent
in the crowd behavior. The Habitat labels assess whether people can recognize the
resulting pedestrian responses in a downstream robot-navigation setting.
The study does not separately measure animation quality or crowd realism.
Responses from the same participant are repeated observations, and the selected
scenes and clips limit the scope of the perceptual conclusions.

\section{Additional Ablation Studies}
\label{app:ablation}
\subsection{Main Component Definitions}
The main component results are reported in Table~\ref{tab:ablation}.
We specify the variants here to distinguish changes to pretraining, planning, and supervision.

\paragraph{Prediction pretraining.}
The randomly initialized variant replaces the prediction-pretrained feature extractor
with a randomly initialized instance of the same architecture while retaining the
subsequent behavior-learning procedure. The full model uses the pretrained, frozen
crowd representation module.
This variant tests the use of prediction-pretrained features.

\paragraph{Planning and learned ranking.}
Actor-only generation bypasses CEM and directly executes actor proposals.
In Table~\ref{tab:ablation}, the full-model generation metrics are computed
with $\lambda_{\mathrm{user}}=0$, while the user-cost-only generation
metrics are computed under avoidance control using $-\lambda_{\mathrm{user}}c_k$
to rank candidates. For the generation metrics, the two variants therefore differ in both the control setting and the ranking rule. The compliance column
compares both variants under the same avoidance protocol.

\paragraph{Training objectives.}
Loss-removal variants retain the actor, critic, and planner.
Collision supervision denotes the collision term in the critic target, rather than
a direct collision penalty in the actor objective.
Generation performance and avoidance compliance use the evaluation windows summarized in
Table~\ref{tab:protocol_summary}.

\subsection{Interaction and Auxiliary Supervision}
These experiments use five ETH--UCY folds, diffusion arrivals, disc avoidance,
$\lambda_{\mathrm{user}}=10$, and 20 trials per scene.
Each disc covers $7.5\%$ of its scene's walkable area, with radii of 18.9--28.5 grid pixels.
Paired runs share the first third, arrivals, and region geometry.
Only the controlled run receives the user cost, and evaluation uses the final two thirds.
This protocol is separate from the 10-trial main avoidance comparison and one-trial sensitivity sweeps.

\paragraph{Interaction encoder.}
Including the interaction encoder increases mean compliance from 0.578 to 0.880,
with improvements on four of five folds (Table~\ref{tab:interaction_ablation}).
Collision effects vary by scene.
Without the encoder, HOTEL, ZARA1, and ZARA2 have fewer collisions but lower compliance.
The large collision increase on ETH accounts for much of the difference in average collision rate.

\paragraph{Auxiliary head.}
The auxiliary head increases mean compliance from 0.584 to 0.880
(Table~\ref{tab:dynamics_ablation}).
Four folds improve, with the largest gains on HOTEL and UNIV.
ZARA1 has higher compliance without the head.
Because the head supplies pretraining supervision and is not used to advance rollout states,
this experiment measures the effect of representation learning rather than adding a transition module.

\subsection{Critic Training Objective}
Table~\ref{tab:component_ablation} compares listwise ranking with pointwise regression under the same 20-trial disc-control protocol. Pointwise regression minimizes the mean squared error between each candidate's critic score and its negative trajectory cost.
Listwise training yields lower average kinematic error, DTW, and collision rate,
and higher compliance.
Its soft targets depend on both the ordering and relative differences of candidate costs.
The collision difference is concentrated in ETH, where the rates are 0.458\% for
listwise training and 7.874\% for pointwise regression.

\subsection{Planning Parameters and Computation}
\label{app:planning_parameters}
Figure~\ref{fig:planning_ablation} varies $K$, $I$, and $H$ individually while retaining
the other defaults in Table~\ref{tab:hparams}.
Motion and compliance metrics are measured over the post-activation window.
The sweep reports latency per planning call, whereas Table~\ref{tab:latency} reports
runtime in milliseconds per simulated second of crowd motion.
The latter measures how long each method takes to generate the same duration of crowd motion;
for Ctrl-CWM, the actor-only setting omits planning.
The 0.5\,s replanning interval in Habitat specifies simulated time between calls,
not the computation time of a call. These quantities should therefore be distinguished
when comparing execution settings.

\section{Trajectory Prediction}
\label{app:prediction}
We evaluate the prediction heads trained jointly with $\phi_{\mathrm{rep}}$ during pretraining.
The task observes eight steps and predicts $L=12$ future steps.
Best-of-20 average displacement error (ADE) and final displacement error (FDE) measure
trajectory-wide and endpoint prediction accuracy, respectively.
ETH--UCY errors are measured in meters, while SDD and GCS errors are in pixels.
AVG is the mean over the five ETH--UCY folds.
The waypoint head uses predicted endpoints during forecasting.
The planner is not used in this forecasting evaluation.

Table~\ref{tab:pred_all} reports the comparison. For AgentFormer and MID, we use the ETH--UCY results reported by Bae et al.~\cite{bae2024can}.
Ctrl-CWM achieves average ETH--UCY ADE/FDE of 0.21/0.31\,m.
GCS errors are 5.89/10.26\,px, displayed as 5.9/10.3\,px in the table.
These results assess the predictive information learned during pretraining.
The generation ablations separately assess its usefulness when the simulator advances from its own states.

\section{Additional Qualitative Results}
\label{app:qualitative}
\begin{figure}[!htbp]
\centering
\includegraphics[width=\linewidth]{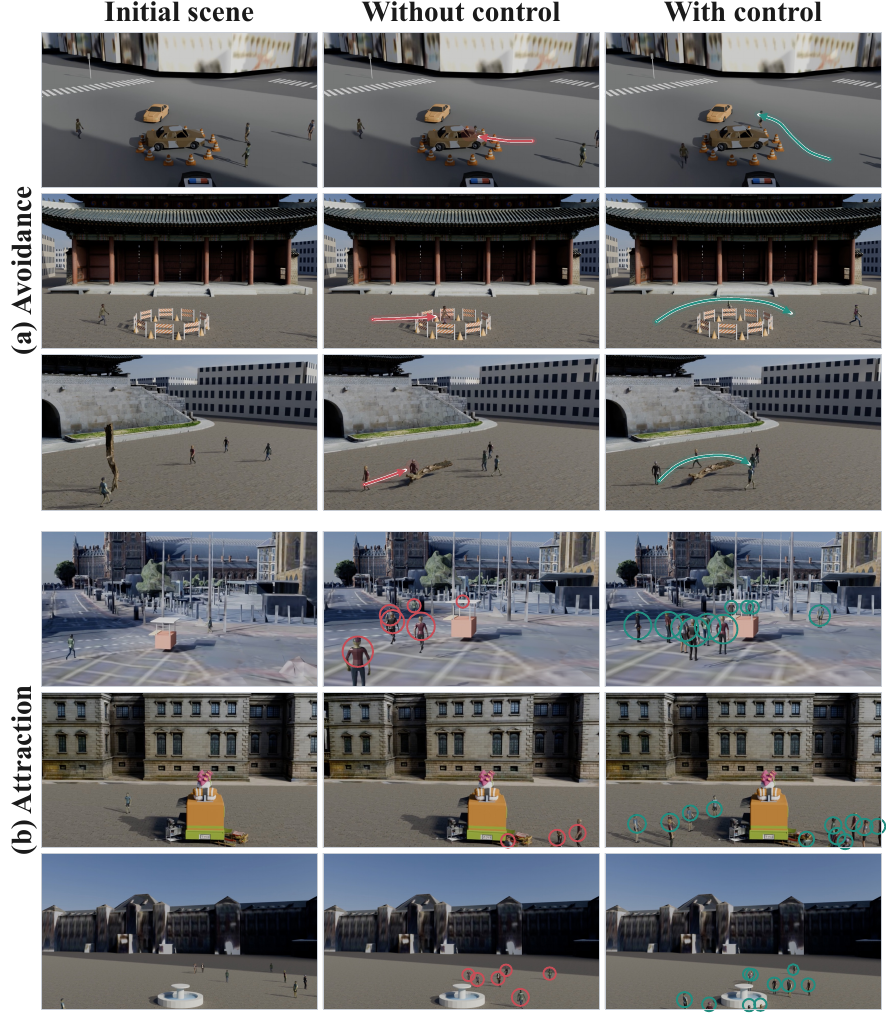}
\caption{\textbf{Run-time control in additional reconstructed scenes.}
Columns show the initial scene, motion without control (red), and motion with control (green).
Each objective is illustrated in three scenes.
(a) Avoidance redirects pedestrians away from a hazardous region.
(b) Attraction guides them toward a specified target.}
\label{fig:runtime_control_extra}
\end{figure}

Figure~\ref{fig:runtime_control_extra} presents three reconstructed scenes for each control objective.
The left column shows the initial scene, while the middle and right columns show
trajectories without and with control.
Avoidance redirects pedestrians away from the designated region, and attraction draws them toward a target.
We visualize the generated coordinate trajectories in CARLA~\cite{dosovitskiy2017carla}
using Unreal Engine~\cite{unrealsoftware}. The humanoid animations provide a visual
realization of trajectory-level control. Improving character animation is outside
the scope of this work. These examples complement the quantitative region-control
results and the moving-robot experiments in Appendix~\ref{app:habitat}.

    \renewcommand{\refname}{Supplementary References}
    {\small\putbib[main]}
\end{bibunit}

\end{document}